\documentclass[10pt,twocolumn,letterpaper]{article}

\usepackage[pagenumbers]{cvpr} 

\usepackage{graphicx}
\usepackage{amsmath}
\usepackage{amssymb}
\usepackage{booktabs}

\usepackage{times}
\usepackage{epsfig}
\usepackage{graphicx} 
\graphicspath{{figure/}}
\usepackage{enumitem}
\usepackage{caption}
\usepackage{dsfont}
\usepackage{wrapfig}
\usepackage{amsfonts,amsmath,amssymb,amsthm}
\usepackage{bm,nicefrac}
\DeclareMathSymbol{\shortminus}{\mathbin}{AMSa}{"39}
\usepackage[bottom]{footmisc}
\usepackage[makeroom]{cancel}
\usepackage[accsupp]{axessibility}  

\newcommand{\bunderbrace}[2]{%
  \begin{array}[t]{@{}c@{}}
  \underbrace{#1}\\
  #2
  \end{array}
}

\newcommand{\duygu}[1]{\textcolor{cyan}{[\textbf{DC}:~#1]}}

\usepackage[pagebackref,breaklinks,colorlinks]{hyperref}

\def\cvprPaperID{2803} 
\def\confName{CVPR}
\def\confYear{2022}

\begin{document}

\title{Learning Motion-Dependent Appearance for High-Fidelity Rendering of Dynamic Humans from a Single Camera}

\author{
Jae Shin Yoon$^\dagger$$^,$$^\sharp$
\hspace{10mm}Duygu Ceylan$^\sharp$
\hspace{10mm}Tuanfeng Y. Wang$^\sharp$
\vspace{1mm}
\\
\hspace{0mm}Jingwan Lu$^\sharp$
\hspace{5mm}Jimei Yang$^\sharp$
\hspace{5mm}Zhixin Shu$^\sharp$
\hspace{5mm}Hyun Soo Park$^\dagger$
\vspace{3mm}
\\
\hspace{-0mm}$^\dagger$University of Minnesota
\hspace{10mm}
$^\sharp$Adobe Research \\
}

\maketitle

\begin{abstract}
Appearance of dressed humans undergoes a complex geometric transformation induced not only by the static pose but also by its dynamics, i.e., there exists a number of cloth geometric configurations given a pose depending on the way it has moved. Such appearance modeling conditioned on motion has been largely neglected in existing human rendering methods, resulting in rendering of physically implausible motion. A key challenge of learning the dynamics of the appearance lies in the requirement of a prohibitively large amount of observations. 
In this paper, we present a compact motion representation by enforcing equivariance---a representation is expected to be transformed in the way that the pose is transformed. We model an equivariant encoder that can generate the generalizable representation from the spatial and temporal derivatives of the 3D body surface. This learned representation is decoded by a compositional multi-task decoder that renders high fidelity time-varying appearance. Our experiments show that our method can generate a temporally coherent video of dynamic humans for unseen body poses and novel views given a single view video.

\end{abstract}

\begin{figure}
	\begin{center}
    \hspace{-5mm}
    \includegraphics[width=1.1\linewidth]{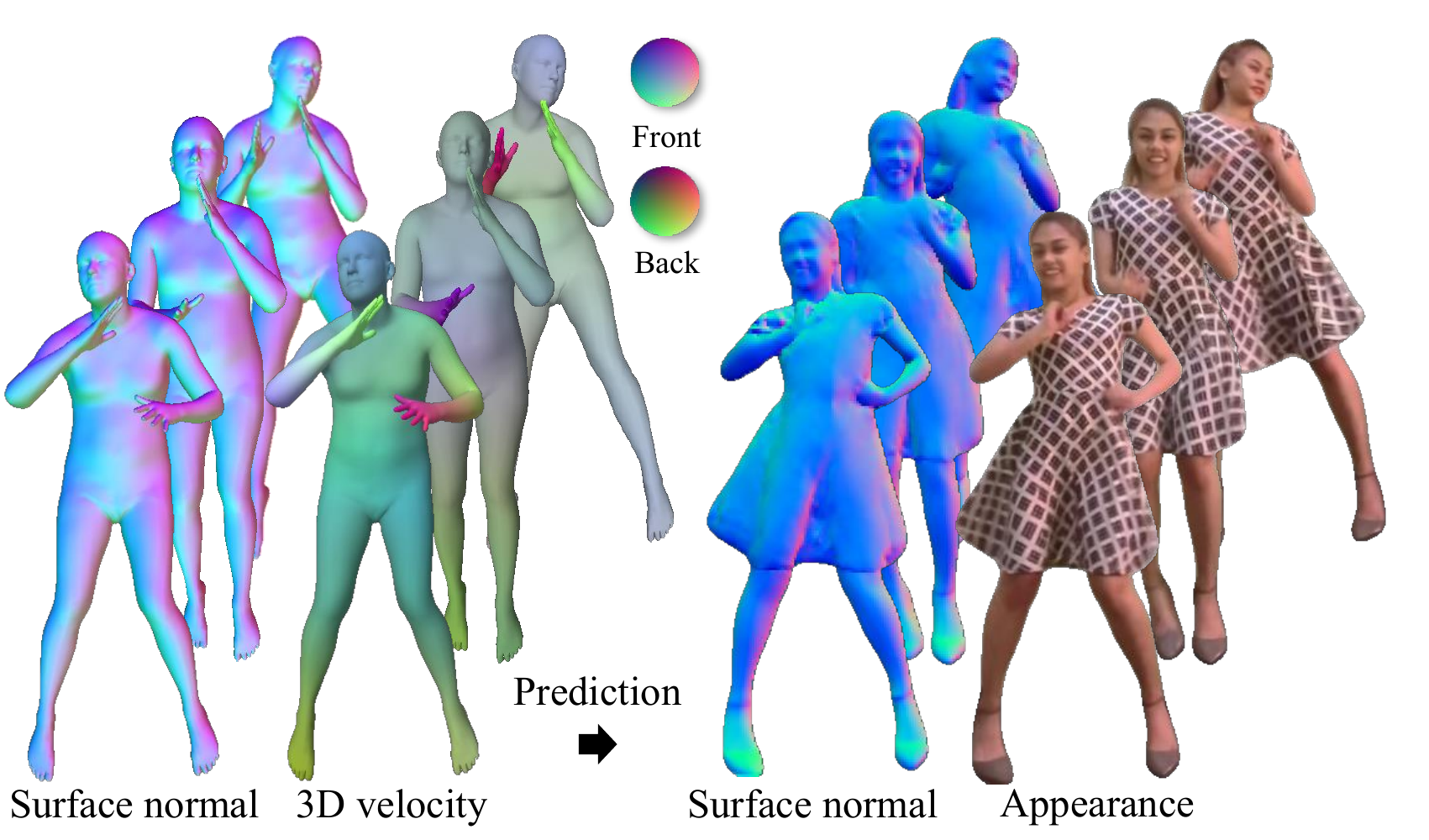}
	\end{center}	
	\vspace{-.5cm}
	\caption{\small Given surface normal and velocity of a 3D body model, our method synthesizes subject-specific surface normal and appearance. We specifically focus on synthesis of plausible dynamic appearance by learning an effective 3D motion descriptor.
	}

	\label{fig:motivation1}
\end{figure}

\section{Introduction}\label{sec:intro}

We express ourselves by moving our body that drives a sequence of natural secondary motion, e.g., dynamic movement of dress induced by dancing as shown in Figure~\ref{fig:motivation1}. This secondary motion is the resultant of complex physical interactions with the body, which is, in general, \textit{time-varying}. This presents a major challenge for plausible rendering of dynamic dressed humans in applications such as video based retargetting or social presence. 
Many existing approaches such as pose-guided person image generation~\cite{chan2019everybody} focus on static poses as a conditional variable. Despite its promising rendering quality, it fails to generate a physically plausible secondary motion, e.g., generating the same appearance for fast and slow motions. 

One can learn the dynamics of the secondary motion from videos. This, however, requires a tremendous amount of data, i.e., videos depicting all possible poses and associated motions. In practice, only a short video clip is available, e.g., the maximum length of videos in social media (e.g., TikTok) are limited to 15-60 seconds. The learned representation is, therefore, prone to overfitting.

In this paper, we address the fundamental question of ``can we learn a representation for dynamics given a limited amount of observations?''. We argue that a meaningful representation can be learned by enforcing an equivariant property---a representation is expected to be transformed in the way that the body pose is transformed. With the equivariance, we model the dynamics of the secondary motion as a function of spatial and time derivative of the 3D body. 
We construct this representation by re-arranging 3D features in the canonical coordinate system of the body surface, i.e., the UV map, which is invariant to the choice of the 3D coordinate system. 

The UV map also captures the semantic meaning of body parts since each body part is represented by a UV patch.
The resulting representation is compact and discriminative compared to the 2D pose representations that often suffer from geometric ambiguity due to 2D projection.

We observe that two dominant factors significantly impact the physicality of the generated appearance. First, the silhouette of dressed humans is transformed according to the body movement and the physical properties (e.g., material) of the individual garment types (e.g., top and bottom garments might undergo different deformations). Second, the local geometry of the body and clothes is highly correlated, e.g., surface normals of T-shirt and body surface, which causes appearance and disappearance of folds and wrinkles. To incorporate these factors, we propose a compositional decoder that breaks down the final appearance rendering into modular subtasks. 
This decoder predicts the time-varying semantic maps and surface normals as intermediate representations. While the semantic maps capture the time-varying silhouette deformations, the surface normals are effective in synthesizing high quality textures, which further enables re-lighting. We combine these intermediate representations to produce the final appearance.

Our experiments show that our method can generate a temporally coherent video of an unseen secondary motion from novel views given a single view training video. We conduct thorough comparisons with various state-of-the-art baseline approaches. Thanks to the discriminative power, our representation demonstrates superior generalization ability, consistently outperforming previous methods when trained on shorter training videos. Furthermore, our method shows better performance in handling complex motion sequences including 3D rotations as well as rendering consistent views in applications such as free-viewpoint rendering. The intermediate representations predicted by our method such as surface normals also enable applications such as relighting which are otherwise not applicable.

\section{Related Work}\label{sec:related}

Two major rendering approaches have been used for high quality human synthesis: model based and retrieval based approaches. Model based approaches leverage the 3D shape models, e.g., 3DMM face model~\cite{Volker99}, that can synthesize a novel view based on the geometric transformation~\cite{Volino:BMVC:2014,Collet:2015}. Retrieval approaches synthesize an image by finding matches in both local and global shape and appearance~\cite{Casas2014,Li2017SPA}. These are, now, combined with a deep representation to form neural rendering and generative networks. 


\begin{figure*}
	\begin{center}
    \includegraphics[width=1\linewidth]{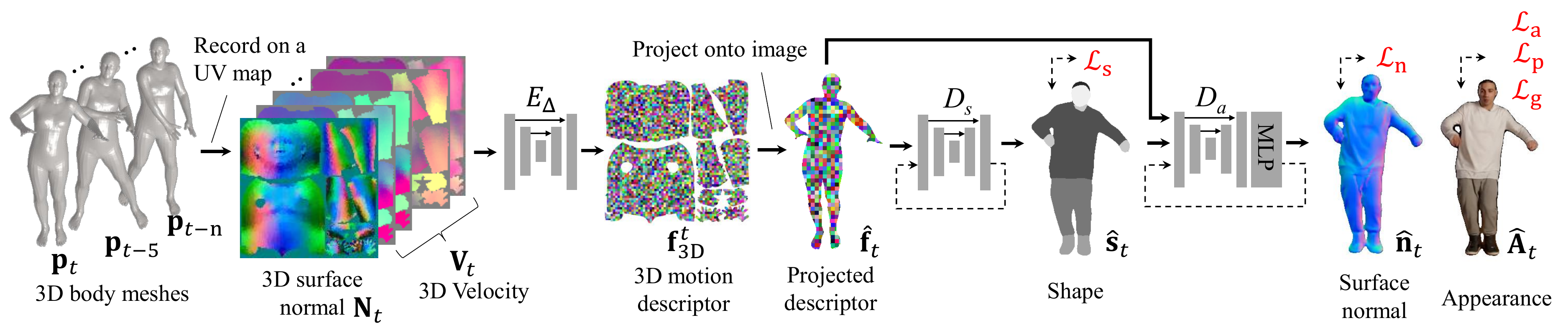}
	\end{center}	
	\vspace{-.4cm}
	\caption{\small The overview of our human rendering pipeline.
Given a set of time-varying 3D body meshes $\{\mathbf{P}_{t}, .. ,\mathbf{P}_{t-n}\}$ obtained from a monocular input video, we aim to synthesize high-fidelity appearance of a dressed human. 
We learn an effective 3D body pose and motion representation by recording the surface normal $\mathbf{N}^{t}$ of the posed 3D mesh at time $t$ and the body surface velocity $\mathbf{V}^{t}$ over several past times in the spatially aligned UV space. 
We define an encoder $E_{\Delta}$ which is designed to reconstruct 3D motion descriptors $\mathbf{f}^{t}_{\rm 3D}$ that encode the spatial and temporal relation of the 3D body meshes. 
Given a target 3D body configuration, we project $\mathbf{f}^{t}_{\rm 3D}$ onto the image space which are then utilized by our compositional networks ($D_s$ and $D_a$) to predict a shape with semantic labels, surface normal, and final appearance. 
%
}
	\vspace{-.4cm}
	\label{fig:rendering}
\end{figure*}

\noindent\textbf{Neural Rendering.} As human appearances are modulated by their poses, it is possible to generate the high fidelity appearance by using a parametric 3D body model, e.g., deformable template models~\cite{SMPL:2015}. For example, some existing approaches~\cite{prokudin2021smplpix, yoon2021pose} learned a constant appearance of a person by mapping from per-vertex RGB colors defined on the SMPL body to synthesized images. Textured neural avatars~\cite{shysheya2019textured} learned a person-specific texture map by projecting the image features to a body surface coordinate (invariant to poses) to model human appearances. These approaches, however, are limited to statics, i.e., the generated appearance is completely blind to pose and motion.


To model pose-dependent appearances, Liu et al.~\cite{liu2019neural} implicitly learned the texture variation over poses, which allowed them to refine the initial appearance obtained from texture map through a template model. On the other hand, Raj et al.~\cite{raj2021anr} explicitly learned pose-dependent neural textures. To further enhance the quality of rendering, person-specific template models~\cite{bagautdinov2021driving} were used by incorporating additional meshes representing the garments~\cite{xiang2021explicit}. However, none of these approaches are capable of modeling the time-varying secondary motion. Habermann et al.~\cite{habermann2021real} utilized a motion cue to model the motion-dependent appearances while requiring a pre-learned person-specific 3D template model. Zhang et al.~\cite{Zhang:2021} proposed a neural rendering approach to synthesize the dynamic appearance of loose garments assuming a coarse 3D garment proxy is provided. In contrast, our method uses the 3D body prior to model the dynamic appearance of both tight and loose garments.


The requirement of the parametric model can be relaxed by leveraging flexible neural rendering fields. For instance, neural volumetric representations~\cite{mildenhall2020nerf} have been used to model general dynamic scenes~\cite{tretschk2020non,xian2021space,gao2021dynamic,wang2021neural,li2021neural} and humans~\cite{park2020deformable,park2021hypernerf} using deformation fields. Nonetheless, the range of generated motion is still limited. Recent methods learned the appearance of a person in the canonical space of the coarse 3D body template and employ skinning and volume rendering approaches to synthesize images~\cite{pumarola2021d,peng2021animatable,chen2021animatable}. Liu et al.~\cite{liu2021neural} extended such approaches by introducing pose-dependent texture maps to model pose-dependent appearance. Modeling time-varying appearance induced by the secondary motion with such volumetric approaches is still the uncharted area of study. 


\noindent\textbf{Generative Networks.} 
Generative adversarial learning enforces a generator to synthesize photorealistic images that are almost indistinguishable from the real images. For example, image-to-image translation can synthesize pose-conditioned appearance of a person by using various pose representations such as 2D keypoints~\cite{ma2017pose,esser2018variational,pumarola2018unsupervised,zhu2019progressive,siarohin2019first,tang2020xinggan,ren2020deep}, semantic labels~\cite{men2020controllable,song2019unsupervised,dong2019towards,balakrishnan2018synthesizing,yang2018pose,han2019clothflow,gafni2021single}, or dense surface coordinate parametrizations~\cite{neverova2018dense,grigorev2019coordinate,sarkarneural,albahar2021pose}. Despite remarkable fidelity, these were built upon the 2D synthesis of static images at every frame, in general, failing to generate physically plausible secondary motion. To address this challenge, several works have utilized temporal cues either in training time~\cite{chan2019everybody} to enable temporally smooth results or as input signals~\cite{wang2018video,wang2019few} to model motion-dependent appearances. Kappel et al.~\cite{kappel2021high} modeled the pose-dependent appearances of loose garments conditioned on 2D keypoints by learning temporal coherence using a recurrent network. However, due to the nature of 2D pose representations, the physicality of the generated motion is limited, e.g., our experiments show that the method works well mostly for planar motions but is limited in expressing 3D rotations. Wang et al.~\cite{wang2021dance} is the closest to our work, which maps a sequence of dense surface parameterization to motion features that are used to synthesize dynamic appearances using StyleGAN~\cite{Karras19}. In contrast, our method is built on a new 3D motion representation which shows superior discriminative power, consistently outperforming~\cite{wang2021dance} in terms of generalizing to unseen poses as demonstrated in our experiments.

\begin{figure*}
	\begin{center}
    \includegraphics[width=1\linewidth]{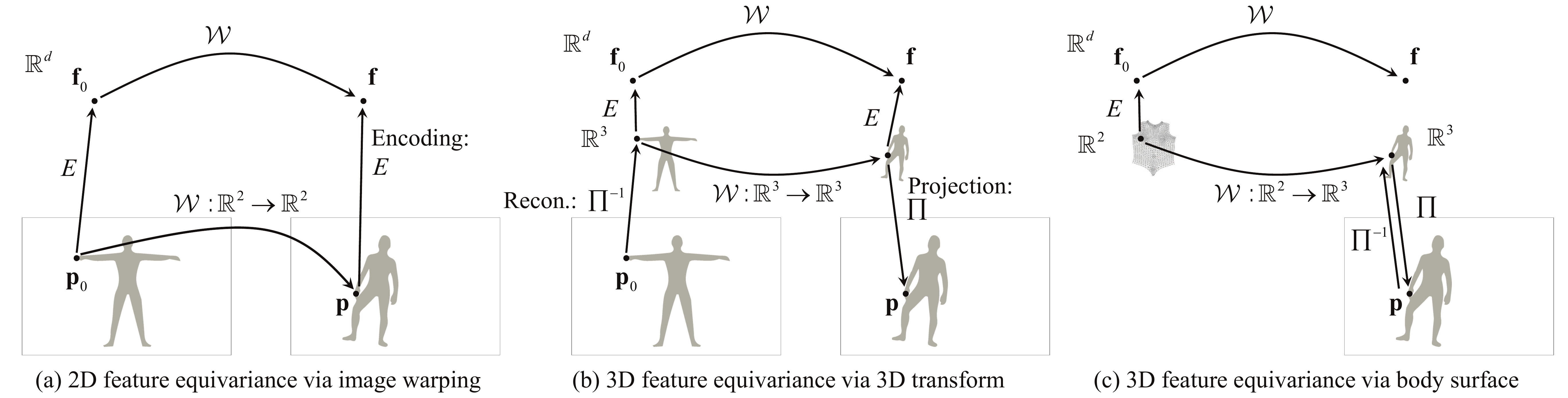}
	\end{center}	
	\vspace{-.4cm}
	\caption{\small We apply equivariance to learn a compact representation. (a) In 2D, the feature $\mathbf{f} = E(\mathbf{p})$ is expected to be transformed to the feature of the neutral pose, $\mathbf{f}_0 = E(\mathcal{W}^{-1}\mathbf{p})$ by a  coordinate transform $\mathcal{W}$, e.g., image warping. This eliminates the necessity of learning the encoder $E$, i.e., the appearance of the pose $\mathbf{p}$ is generated by warping the appearance of the neutral pose. (b) Equivariance in 3D can be applied by incorporating 3D body reconstruction $\Pi^{-1}$ where the feature is expected to be transformed by the 3D warping $\mathcal{W}$, e.g., skinning. (c) We use the canonical body surface coordinate (UV coordinate) to represent the feature coordinate transformation. 
	}
	\vspace{-.4cm}
	\label{fig:equivariance}
\end{figure*}

\section{Method}\label{sec:method}
Given a monocular video of a person in motion and the corresponding 3D body fit estimates, we learn a motion representation to describe the time-varying appearance of the secondary motion induced by body movement (Section~\ref{subsec:spamod}).
We propose a multitask compositional 
renderer (Section~\ref{sec:renderer}) that uses this representation to render the subject-specific final appearance of moving dressed humans. Our renderer first predicts two intermediate representations including time-varying semantic maps that capture garment specific silhouette deformations and surface normals that capture the local geometric changes such as folds and wrinkles. These intermediate representations are combined to synthesize the final appearance. We obtain the 3D body fits estimates from the input video using a new model-based tracking optimization (Supplementary material). The overview of our rendering framework is shown in Figure~\ref{fig:rendering}.


%
%
%

\subsection{Equivariant 3D Motion Descriptor}\label{subsec:spamod}
We cast the problem of human rendering as learning a representation via a feature encoder-decoder framework:
\begin{align}
    \mathbf{f} = E({\mathbf{p}}),\ \ \  \mathbf{A}=D(\mathbf{f})\label{eq1_1},
\end{align}
where an encoder $E$ takes as an input a representation of a posed body, $\mathbf{p}$ (e.g., 2D sparse or dense keypoints or 3D body surface vertices), and outputs per-pixel features $\mathbf{f}$ 
that can be used by the decoder $D$ to reconstruct the appearance $\mathbf{A}\in [0,1]^{w\times h\times 3}$ of the corresponding pose where $w$ and $h$ are the width and height of the output image (appearance). We first discuss how $E$ can be modeled to render static appearance, then introduce our 3D motion descriptor to render time-varying appearance with secondary motion effects.

Learning a representation from Equation~(\ref{eq1_1}) from a limited amount of data is challenging because both encoder and decoder need to memorize every appearance in relation to the corresponding pose, $\mathbf{A}\leftrightarrow\mathbf{p}$. To address the data challenge, one can use an \textit{equivariant} geometric transformation, $\mathcal{W}$, such that a feature is expected to be transformed in the way that the body pose is transformed:
\begin{align}
    E(\mathcal{W}\mathbf{x})=\mathcal{W}E(\mathbf{x}).
    \label{eq:equivariance}
\end{align}
where $\mathbf{x}$ is an arbitrary pose.
A naive encoder that satisfies this equiavariance learns a constant feature $\mathbf{f}_0$ by warping any $\mathbf{p}$ to a neutral pose $\mathbf{p}_0$:
%
\begin{align}
    \mathbf{f}_0 = E(\mathcal{W}^{-1}\mathbf{p}) = {\rm const.}, \ \ \ \mathbf{A} = D(\mathcal{W}\mathbf{f}_0), \label{Eq:image_warp}
\end{align}
where $\mathbf{p} = \mathcal{W}\mathbf{p}_0$. 
Figure~\ref{fig:equivariance}(a) and (b) illustrate cases where $\mathcal{W}$ is defined as image warping in 2D or skinning in 3D respectively.  $\mathbf{f}$ can be derived by warping $\mathbf{p}$ to the T-pose, $\mathcal{W}^{-1}\mathbf{p}$ of which feature can be warped back to the posed feature before decoding, $D(\mathcal{W}E(\mathcal{W}^{-1}\mathbf{p}))$. Since $\mathbf{f}_0$ is constant, the encoder $E$ does not need to be learned. One can only learn the decoder $D$ to render a static appearance.

To model the time-varying appearance for the secondary motion that depends on both body pose and motion, one can extend Equation~(\ref{Eq:image_warp}) to encode the spatial and temporal gradients as a residual feature encoding:
\begin{align}
    &\mathbf{f}=E\left({\mathbf{p}},\frac{\partial {\mathbf{p}}}{\partial{x}}, \frac{\partial {\mathbf{p}}}{\partial{t}}     \right)
    \approx E({\mathbf{p}}) + E_{\Delta}\left(\frac{\partial {\mathbf{p}}}{\partial{x}}, \frac{\partial {\mathbf{p}}}{\partial{t}}\right)\nonumber\\
    \Longleftrightarrow~&\mathbf{f}_0 = \bunderbrace{\cancel{E(\mathcal{W}^{-1}{\mathbf{p}})}}{\rm const.} + E_{\Delta}\left(\mathcal{W}^{-1}\frac{\partial {\mathbf{p}}}{\partial{x}}, \mathcal{W}^{-1}\frac{\partial {\mathbf{p}}}{\partial{t}}\right),\label{eq1_2}
\end{align}
where $\frac{\partial}{\partial{x}}$ and $\frac{\partial}{\partial{t}}$ are the spatial and temporal derivatives of the posed body, respectively. 
The spatial derivatives essentially represent the pose corrective deformations~\cite{Lewis2000,SMPL:2015}. The temporal derivatives denote the body surface velocity which results in secondary motion. Since these spatial and temporal gradients are no longer constant, one needs to learn an encoder $E_{\Delta}$ to encode the residual features.


\begin{figure*}
	\begin{center}
    \includegraphics[width=1\linewidth]{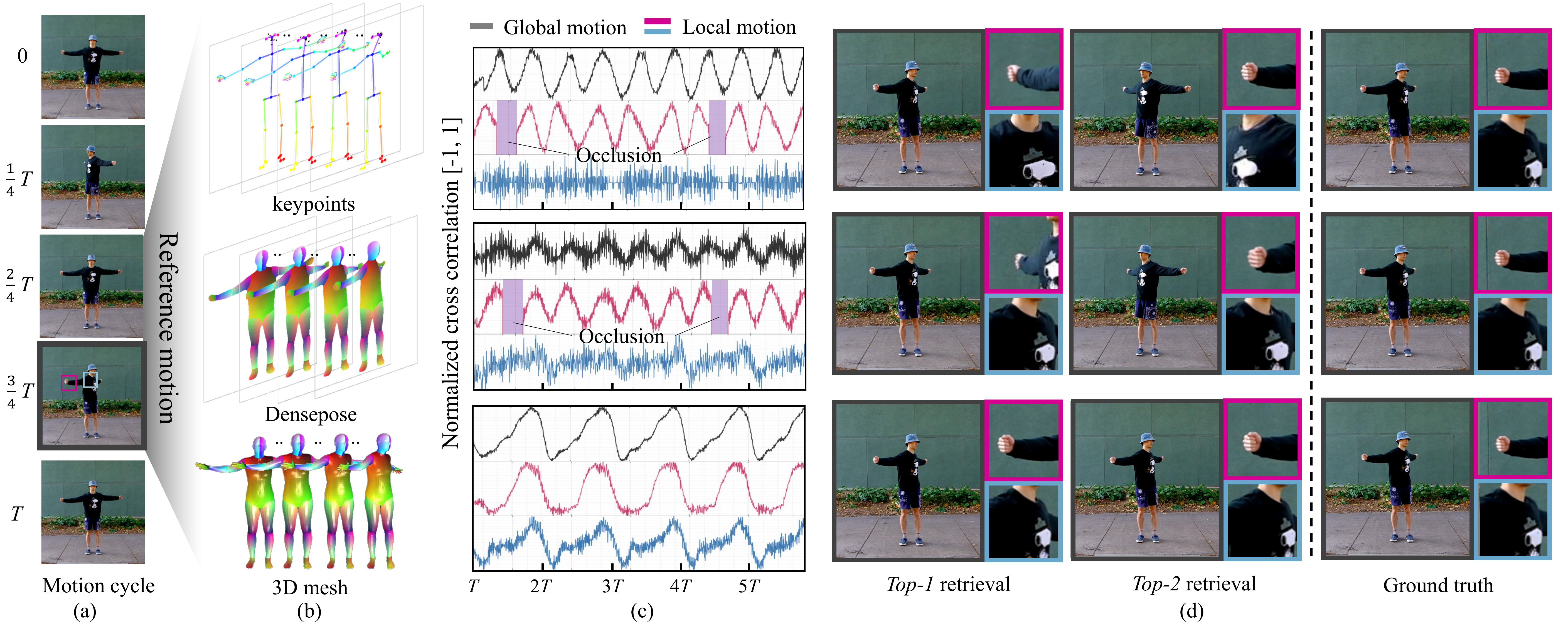}
	\end{center}	
	\vspace{-.4cm}
	\caption{\small We show the strength of our 3D motion descriptor using a toy example. Given a video of a person rotating his body from left to right multiple times, we associate the first cycle of the motion (\textit{i.e.,} $0\sim T$) to the remaining cycles ($T\sim6T$). As a proof-of-concept, we use a nearest neighbor classifier to model $D$.
	%
	(b) We represent the motion descriptor using (top) 2D keypoints~\cite{cao2017realtime}, (middle) 2D dense UV coordinates~\cite{guler2018densepose}, (bottom) and 3D body mesh~\cite{SMPL:2015}.
	%
	(c) We measure the similarity in motion descriptor for entire body (gray), local hand (pink) and upper torso (blue) using normalized cross correlation (NCC) where multiple peaks within a cycle indicate ambiguity of the descriptor. 
	%
    (d) Given the motion descriptors, we retrieve relevant image patches. 
    While the 3D motion descriptors identify the image patches similar to the ground truth, due to the depth ambiguity, the 2D motion descriptors result in ambiguous matches. Furthermore, the 2D motion descriptors are not well defined in case of occlusions.
	}
	\vspace{-.4cm}
	\label{fig:motivation}
\end{figure*}

In this paper, we use a 3D representation of the posed body lifted from an image by leveraging recent success in single view pose reconstruction~\cite{kanazawa2018end,choutas2020monocular,kocabas2020vibe}. Hence, spatial and temporal derivatives of the body pose correspond to the surface normals and body surface velocities, respectively: 
\begin{align}
    \mathbf{f}_{\rm 3D}=E_{\Delta}(\mathcal{W}^{-1}\mathbf{N},\mathcal{W}^{-1}\mathbf{V}), \ \ \mathbf{A} = D(\Pi \mathcal{W}\mathbf{f}_{\rm 3D}),\label{eq1_4}
\end{align}
where $\mathbf{N}=\frac{\partial {\mathbf{p}}}{\partial{X}}\in\mathds{R}^{m\times3}$ is the 3D surface normal, and $\mathbf{V}=\frac{\partial {\mathbf{p}}}{\partial{t}}\in\mathds{R}^{m\times3}$ represents the instantaneous velocities of the $m$ vertices in the body surface. 
%
We model the geometric transformation function $\mathcal{W}$ to warp an arbitrary 3D pose $\mathbf{p}$ to a canonical representation, $\mathbf{p}_0$. We record $\textbf{f}_{\rm 3D}$ in a spatially aligned 2D positional map, specifically the UV map of the 3D body mesh where each pixel contains the 3D information of a unique point on the body mesh surface. This enables us to leverage 2D CNNs to apply local convolution operations to capture the relationship between neighboring body parts~\cite{ma2021scale}. 
Therefore, $\mathbf{f}_{\rm 3D}\in \mathds{R}^{m\times d}$ called \textit{3D motion descriptor} is the feature defined in the UV coordinates where $d$ is the dimension of the per-vertex 3D feature. $\mathbf{f}=\Pi\mathcal{W}\mathbf{f}_{\rm 3D}$ is the projected 3D feature in the image coordinates where $\Pi$ is a coordinate transformation that transports the features defined in the UV space to the image plane via the dense UV coordinates of the body mesh.

The key advantage of the 3D motion descriptor over commonly used 2D sparse~\cite{kappel2021high} or dense~\cite{wang2021dance} keypoint representations is discriminativity. 
Consider a toy example of a person rotating his body left to right multiple times. Given one cycle (i.e., 0-\textit{T}) of such a motion as input (Figure~\ref{fig:motivation}(a)), assume we want to synthesize the appearance of the person performing the repetitions of the same motion (i.e., cycles \textit{T}-5\textit{T}). As a proof of concept, we model $D$ using a nearest neighbor classifier to retrieve the relevant image patches (top two patches) for each body part from the reference motion based on the correlation of the motion descriptors as shown in Figure~\ref{fig:motivation}(c). Due to the inherent depth ambiguity, multiple 3D motion trajectories yield the same 2D projected trajectory~\cite{JainSCA10}. Hence, the 2D motion descriptors using sparse (Figure~\ref{fig:motivation}(b), top) and dense (Figure~\ref{fig:motivation}(b), middle) 2D keypoints confuse the directions of out-of-plane body rotation, resulting in erroneous nearest neighbor retrievals as shown in Figure~\ref{fig:motivation}(d). Furthermore, 2D representations entangle the viewpoint and pose into a common feature. This not only avoids a compact representation (e.g., the same body motion maps to different 2D trajectories with respect to different viewpoints and yields different features) but also suffers from occlusions in the image space. In our example in Figure~\ref{fig:motivation}(c), top and bottom, the upper arm is occluded in portions of the input video denoted with the purple blocks, hence no reliable local motion descriptors can be computed at those time instances. In contrast, our 3D motion descriptor is highly discriminative, which does not confuse the direction of body rotations, resulting in accurate image patch retrievals. 


\subsection{Multitask Compositional Decoder}
\label{sec:renderer}
Given 3D motion features, the decoder $D$ still needs to learn to generate diverse and plausible secondary motion, which is prone to overfitting given a limited amount of data. We integrate the following properties that can mitigate this challenge. (1) Composition: we design the decoder using a composition of modular functions where each modular function is learned to generate physically and semantically meaningful intermediate representations. Learning each modular function is more accurate than learning an end-to-end decoder as a whole as shown in our ablation study (Section~\ref{sec:eval}); (2) Multi-task: each intermediate representation receives its own supervision signals resulting in multi-task learning. The motion features, $\mathbf{f}_{\rm 3D}$, are shared by all intermediate modules resulting in a compact representation; (3) Recurrence: each module is modeled as an autoregressive network, which allows learning the dynamics rather than memorizing the pose-specific appearance. 

Our decoder is a composition of two modular functions:
\begin{align}
    D = D_a \circ D_s,
\end{align}
where $D_s$ and $D_a$ are the functions that generate the shape with semantic maps and the appearance. 

$D_s$ learns the dynamics of the 2D shape:
\begin{align}
    \widehat{\mathbf{s}}_t = D_s (\widehat{\mathbf{s}}_{t-1}; \widehat{\mathbf{f}}_{t})
\end{align}
where $\widehat{\mathbf{f}}_t = \Pi\mathcal{W}_t \mathbf{f}_{\rm 3D}^t$ is the projected features onto the image at time $t$, and $\widehat{\mathbf{s}_t} \in \{0,\cdots,L\}^{w\times h}$ is the predicted shape with semantics where $L$ is the number of semantic categories. In our experiments we set $L=7$ (background, top clothing, bottom clothing, face, hair, skin, shoes).

$D_a$ learns the dynamics of appearance given the shape and 3D motion descriptor:
\begin{align}
    \widehat{\mathbf{A}}_t, \widehat{\mathbf{n}_t} = D_a(\widehat{\mathbf{A}}_{t-1}, \widehat{\mathbf{n}}_{t-1}; \widehat{\mathbf{s}}_t, \widehat{\mathbf{f}}_{t}),
\end{align}
where $\widehat{\mathbf{A}}_t \in \mathds{R}^{w\times h \times 3}$ and $\widehat{\mathbf{n}_t}\in \mathds{R}^{w\times h\times 3}$ are the generated appearance and surface normals at time $t$. 

We learn the 3D motion descriptor as well as the modular decoder functions by minimizing the following loss:
\begin{align}
\mathcal{L} = \sum_{\mathbf{P}, \mathbf{A} \in \mathcal{D}} \mathcal{L}_{\rm a} + \lambda_{\rm s}\mathcal{L}_{\rm s} + \lambda_{\rm n}\mathcal{L}_{\rm n} + \lambda_{\rm p}\mathcal{L}_{\rm p}+\lambda_{\rm g}\mathcal{L}_{\rm g},\label{eq2_4}
\end{align}
where $\mathcal{L}_{\rm a}$, $\mathcal{L}_{\rm s}$, $\mathcal{L}_{\rm n}$, $\mathcal{L}_{\rm p}$, $\mathcal{L}_{\rm g}$ are the appearance, shape, surface normal, perceptual similarity, and generative adversarial losses, and $\lambda_{\rm s}$, $\lambda_{\rm n}$, $\lambda_{\rm p}$, and $\lambda_{\rm g}$ are their weights, respectively. We set $\lambda_{\rm s}=10$, $\lambda_{\rm n}=\lambda_{\rm p}=1$ and $\lambda_{\rm g}=0.01$ in our experiments. $\mathcal{D}$ is the training dataset composed of the ground truth 3D pose $\mathbf{P}$ and its appearance $\mathbf{A}$. 
\begin{align}
    \mathcal{L}_{\rm a}(\mathbf{P},\mathbf{A}) &= \|\widehat{\mathbf{A}} - \mathbf{A}\|,\nonumber\\
    \mathcal{L}_{\rm s}(\mathbf{P},\mathbf{A}) &= \|\widehat{\mathbf{s}} - S(\mathbf{A})\|,\nonumber\\
    \mathcal{L}_{\rm n}(\mathbf{P},\mathbf{A}) &= \|\widehat{\mathbf{n}} - N(\mathbf{A})\|,\nonumber\\
    \mathcal{L}_{\rm p}(\mathbf{P},\mathbf{A}) &= \sum_i \|VGG_i(\widehat{\mathbf{A}}) - VGG_i(\mathbf{A})\|, \nonumber\\
    \mathcal{L}_{\rm g}(\mathbf{P},\mathbf{A}) &= \mathds{E}_{S(\mathbf{A}),\mathbf{A}}[{\rm log}({D}^{\star}(S(\mathbf{A}), \mathbf{A})]+ \nonumber\\
& ~~~~~\mathds{E}_{S(\mathbf{A}),\widehat{\mathbf{A}}}[{\rm log}(1-D^{*}(S(\mathbf{A}), \widehat{\mathbf{A}})], \nonumber
\end{align}
where $\widehat{\mathbf{A}}$, $\widehat{\mathbf{s}}$, and $\widehat{\mathbf{n}}$ are the generated appearance, shape, and surface normal, respectively. $S$ and $N$ are the shape~\cite{gong2018instance} and surface normal estimates~\cite{Jafarian_2021_CVPR_TikTok}, and $VGG$ is the feature extractor that computes perceptual features from conv-$i$-2 layers in VGG-16 networks~\cite{Simonyan15}, $D^{*}$ is the PatchGAN discriminator~\cite{isola2017image} that validates the plausibility of the synthesized image conditioned on the shape mask. 

\begin{figure*}[t]
	\begin{center}
    \includegraphics[width=1.02\linewidth]{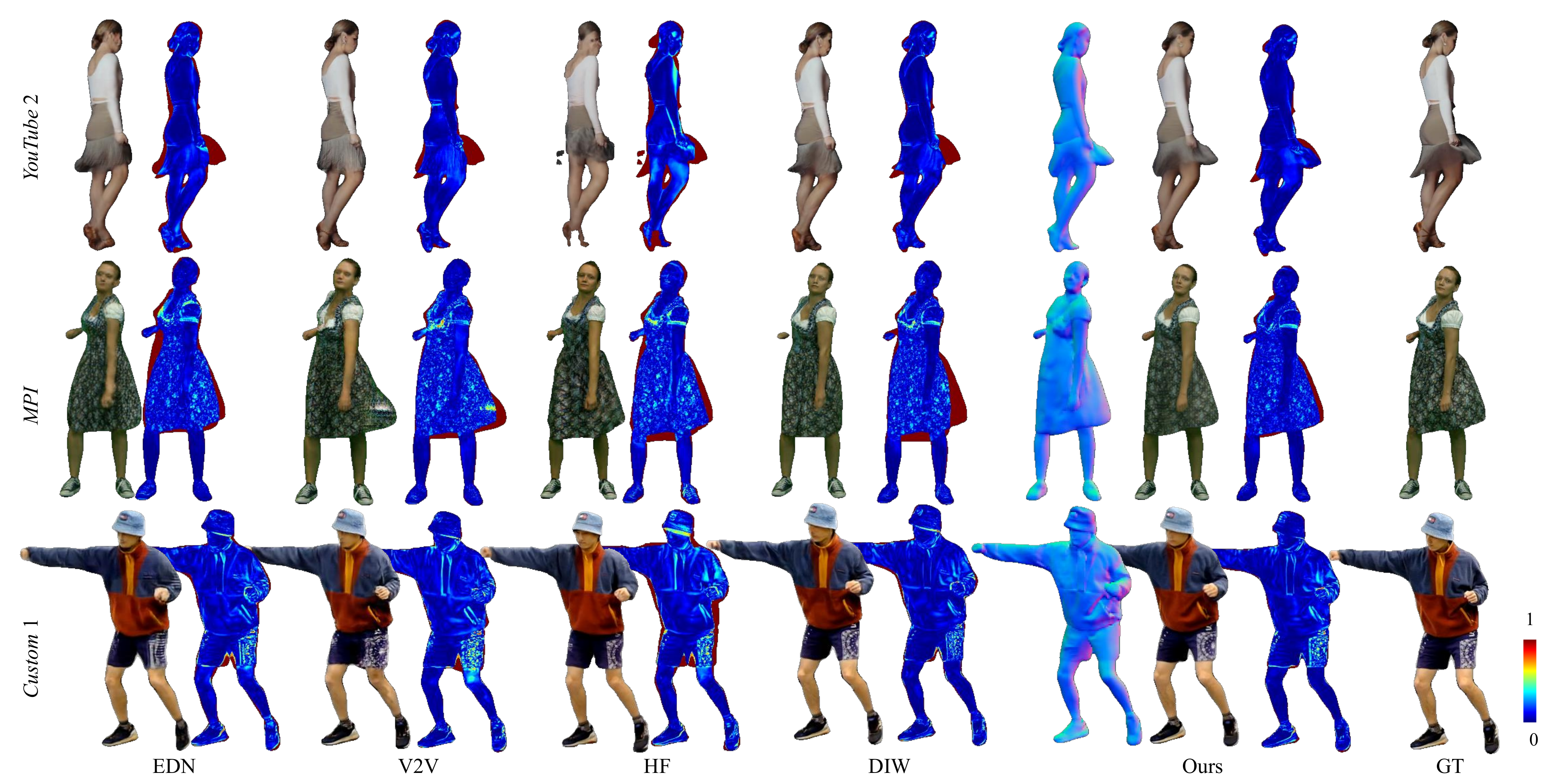}
	\end{center}	
	\vspace{-.6cm}
	\caption{\small We compare our method to several baselines (EDN~\cite{chan2019everybody}, V2V~\cite{wang2018high}, HFMT~\cite{kappel2021high}, DIW~\cite{wang2021dance}) on various sequences. For each example, we show the ground truth (GT) target appearance, the synthesized appearance by each method, and a color map of the error between the two. For our method, we also visualize the predicted surface normal.}
	\label{fig:render_res}
\end{figure*}

\subsection{Model-based Monocular 3D Pose Tracking}\label{method:perform}
While there has been significant improvements in monocular 3D body estimation~\cite{kocabas2020vibe,kolotouros2019learning,choutas2020monocular}, we observe that predicting accurate and temporally coherent 3D body sequences is still challenging, which inhibit to reconstruct high-quality 3D motion descriptors. Hence, we devise a new optimization framework that learns a tracking function to address this challenge. We describe the full methodology and evaluation of our 3D tracking pipeline in the supplementary materials.

{
\renewcommand{\tabcolsep}{5pt}
\begin{table*}[t]
\centering
\scriptsize
\begin{tabular}{l|cccccc|c}
\toprule
  Method & \scriptsize{{\textit{YouTube} 1} (6K)}  & \scriptsize{{\textit{YouTube} 2} (10K)}  & \scriptsize{\textit{YouTube }3 (4K)}  & \scriptsize{{\textit{MPI}} (10K)} & \scriptsize{{\textit{Custom} 1} (15K)}  & \scriptsize{{\textit{Custom} 2} (15K)} & \scriptsize{Avg.} \\
\hline
\scriptsize{EDN}~\cite{chan2019everybody} &\scriptsize{0.954 / 3.06 / 0.356} &\scriptsize{0.943 / 4.39 / 0.465} &\scriptsize{0.871 / 6.23 / 0.467}   &\scriptsize{0.824 / 4.59 / 0.287} &\scriptsize{0.916 / 5.26 / 0.450}&  \scriptsize{0.928 / 5.06 / 0.423}&\scriptsize{0.906 / 4.76 / 0.408}      \\
\scriptsize{V2V}~\cite{wang2018high} &\scriptsize{0.960 / {2.23}  / {\color{red}0.235}} &\scriptsize{0.958 / 3.33 / {0.405}} &\scriptsize{0.880 / {\color{red}4.47} / {\color{blue}0.401}}   &\scriptsize{0.824 / 3.58 / 0.298} &\scriptsize{0.935 / 3.52 / 0.306}&  \scriptsize{{0.943} / 4.15 / {\color{red}0.385}}&\scriptsize{0.916 / 3.54 / 0.338}      \\
\scriptsize{HFMT}~\cite{kappel2021high} &\scriptsize{0.944 / 4.19 / 0.412} &\scriptsize{0.923 / 6.63 / 0.775} &\scriptsize{0.862 / 7.16 / 0.456}   &\scriptsize{{\color{red}0.826} / 5.03 / {\color{blue}0.291}} &\scriptsize{0.905 / 6.24 / 0.321}&  \scriptsize{0.915 / 6.63 / {\color{blue}0.390}}&\scriptsize{0.895 / 5.98 / 0.440}      \\
\scriptsize{DIW}~\cite{wang2021dance} &\scriptsize{{\color{blue}0.966} / {\color{blue}2.21} / 0.275} &\scriptsize{{\color{blue}0.960} / {\color{blue}3.03} / {\color{blue}0.370}} &\scriptsize{{\color{blue}0.894} / 4.69  / {\color{red}0.396}}   &\scriptsize{{\color{blue}0.825} / {\color{blue}2.94} / 0.359} &\scriptsize{{\color{blue}0.939} / {\color{blue}3.23} / {\color{blue}0.304}}&  \scriptsize{{\color{blue}0.944} / {\color{blue}3.95} / 0.412}&\scriptsize{{\color{blue}0.921} / {\color{blue}3.34} / {\color{blue}0.336}}      \\
\hline
\scriptsize{Ours} &\scriptsize{{\color{red}0.973} / {\color{red}2.01} / {\color{blue}0.240} } &\scriptsize{{\color{red}0.964} / {\color{red}2.83} / {\color{red}0.338}} &\scriptsize{{\color{red}0.897} / {\color{blue}4.50} / 0.412}   &\scriptsize{0.825 / {\color{red}2.82} / {\color{red}0.203}} &\scriptsize{{\color{red}0.942} / {\color{red}3.12} / {\color{red}0.279}}&  \scriptsize{{\color{red}0.946} / {\color{red}3.81} / 0.404}&\scriptsize{{\color{red}0.925} / {\color{red}3.18} / {\color{red}0.312}}      \\

\bottomrule
\end{tabular}
\vspace{-1mm}
\caption{Quantitative results. The number of training frames in each sequence is given in the top row. The thee numbers are the SSIM ($\uparrow$), LPIPS ($\downarrow$)$\times$100, and tLPIPS ($\downarrow$)$\times$100 metrics, respectively. The {\color{red}red} represents the best performer, and the {\color{blue}blue} second best.}
\label{table:full_res}
\end{table*}
}

\section{Experiments}\label{sec:exp}
We validate the performance of our method across various examples and perform extensive qualitative and quantitative comparisons with previous work.

\noindent\textbf{Implementation Details.}\label{method:inference}
We utilize the Adam optimizer~\cite{kingma2013auto} to train our model with a learning rate of $1\times10^{-3}$. Given an input video ($\sim 10K$ frames), we train our model for roughly 72 hours using 4 NVIDIA V100 GPUs using a batch size of 4. 
Our motion features are learned from the body surface normals in the current frame and the body surface velocities in the past $t=10$ frames. The features are recorded in a UV map of size 128$\times$128. We synthesize final renderings and surface normal maps of size 512$\times$512. 
We implement our model in Pytoch and utilize the Pytorch3D differentiable rendering layers~\cite{ravi2020pytorch3d}. Details of the network designs and 3D tracking pipeline are given in the supplementary materials.

The coordinate transformation that transports the motion features from the UV space to the image plane, i.e., $\Pi$ in Equation~(\ref{eq1_4}), can be implemented either by using image-based dense UV estimates~\cite{guler2018densepose} or by directly rendering the UV coordinates of the 3D body fits. To provide a fair comparison with previous work which also utilize dense UV estimates, we use the former option. When demonstrating our method on applications where we do not have corresponding ground truth frames to estimate dense UV maps (e.g., novel viewpoint synthesis), we use the latter option. 

%
%
%
\noindent\textbf{Baselines.} We compare ours to four prior methods that focused on synthesizing dressed humans in motion. 1) \underline{Everybody dance now (EDN)}~\cite{chan2019everybody} uses image-to-image translation to synthesize human appearance conditioned on 2D keypoints and uses a temporal discriminator to enforce plausible dynamic appearance. 2) \underline{Video-to-video translation (V2V)}~\cite{wang2018high} is a sequential video generator that synthesizes high-quality human renderings from 2D keypoints and dense UV maps where the motion is modeled with optical flow in the image space. 3) \underline{High-fidelity motion transfer (HFMT)}~\cite{kappel2021high} is a compositional recurrent network which predicts plausible motion-dependent shape and appearance from 2D keypoints. 4) \underline{Dance in the wild (DIW)}~\cite{wang2021dance} synthesizes dynamic appearance of humans based on a motion descriptor composed of time-consecutive 2D keypoints and dense UV maps. We evaluate only on foreground by removing the background synthesized by the methods EDN, V2V, and DIW using a human segmentation method~\cite{gong2018instance}. HFMT predicts a foreground mask similar to ours. 
In the supplementary material, we also compare our method to the 3D based approach~\cite{chen2021animatable} for neural avatar modeling from a single camera, which explicitly reconstruct the geometry of animatable human.

\noindent\textbf{Datasets.} We perform experiments on video sequences that demonstrate a wide range of motion sequences and clothing types, which include non-trivial secondary motion. Specifically, we select three dance videos (e.g., hip-hop and salsa) from YouTube and one sequence from prior work~\cite{kappel2021high} that shows a female subject in a large dress. We also capture two custom sequences showing a male and a female subject respectively performing assorted motions (e.g., walking, running, punching, jumping etc.) including 3D rotations. 

\noindent\textbf{Metrics.} We measure the quality of the synthesized frames with two metrics: 1) Structure similarity (SSIM)~\cite{wang2004image} compares the local patterns of pixel intensity in the normalized luminance and contrast space. 2) Perceptual distance (LPIPS)~\cite{zhang2018unreasonable} evaluates the cognitive similarity of a synthesized image to ground truth by comparing their perceptual features extracted from a deep neural network. We evaluate the temporal plausibility by comparing the perceptual change across frames~\cite{chu2020learning}: ${\rm tLPIPS}=\|{\rm LPIPS}(s_t,s_{t-1})-{\rm LPIPS}(g_t,g_{t-1})\|$ where $s$ and $g$ are the synthesized and ground truth images. 

\begin{figure}
	\begin{center}
    \includegraphics[width=1\linewidth]{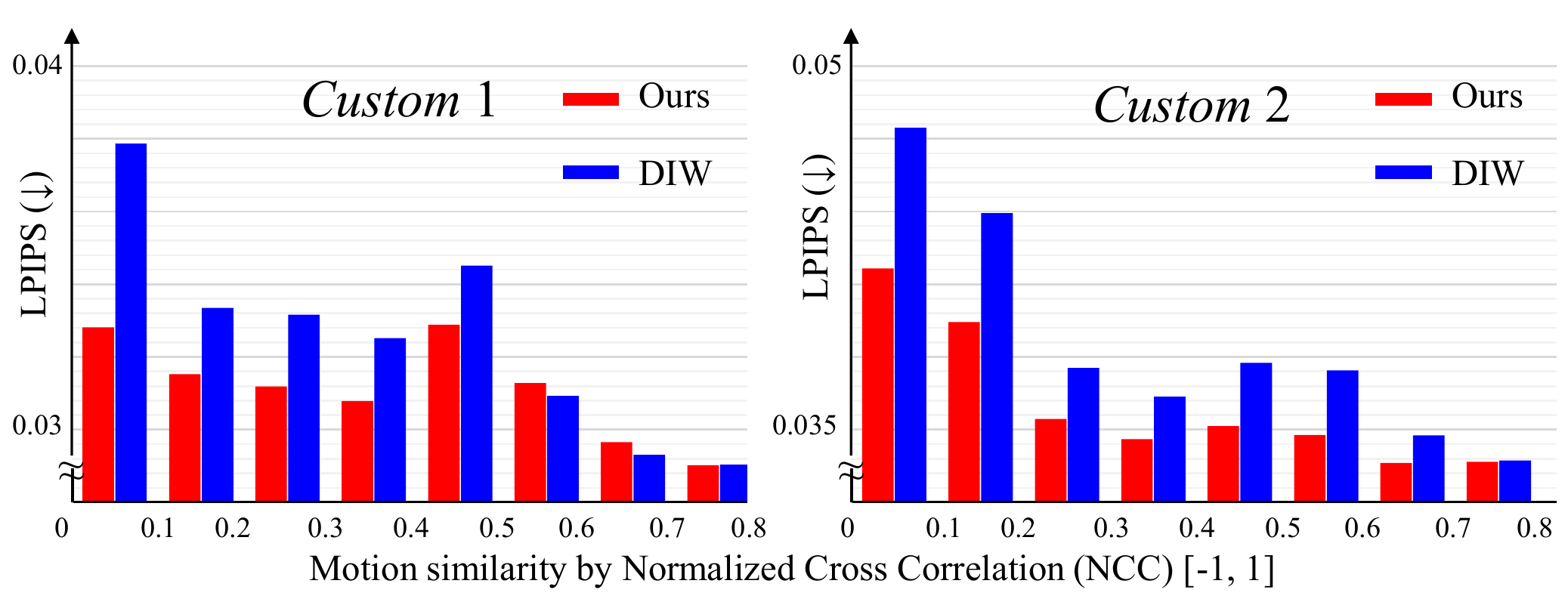}
	\end{center}	
	\vspace{-.6cm}
	\caption{\small Perceptual quality of a synthesized image over motion similarity between training and testing sequences.}
	\vspace{-.4cm}
	\label{fig:generalization_graph}
\end{figure}
\begin{figure}
	\begin{center}
    \includegraphics[width=1\linewidth]{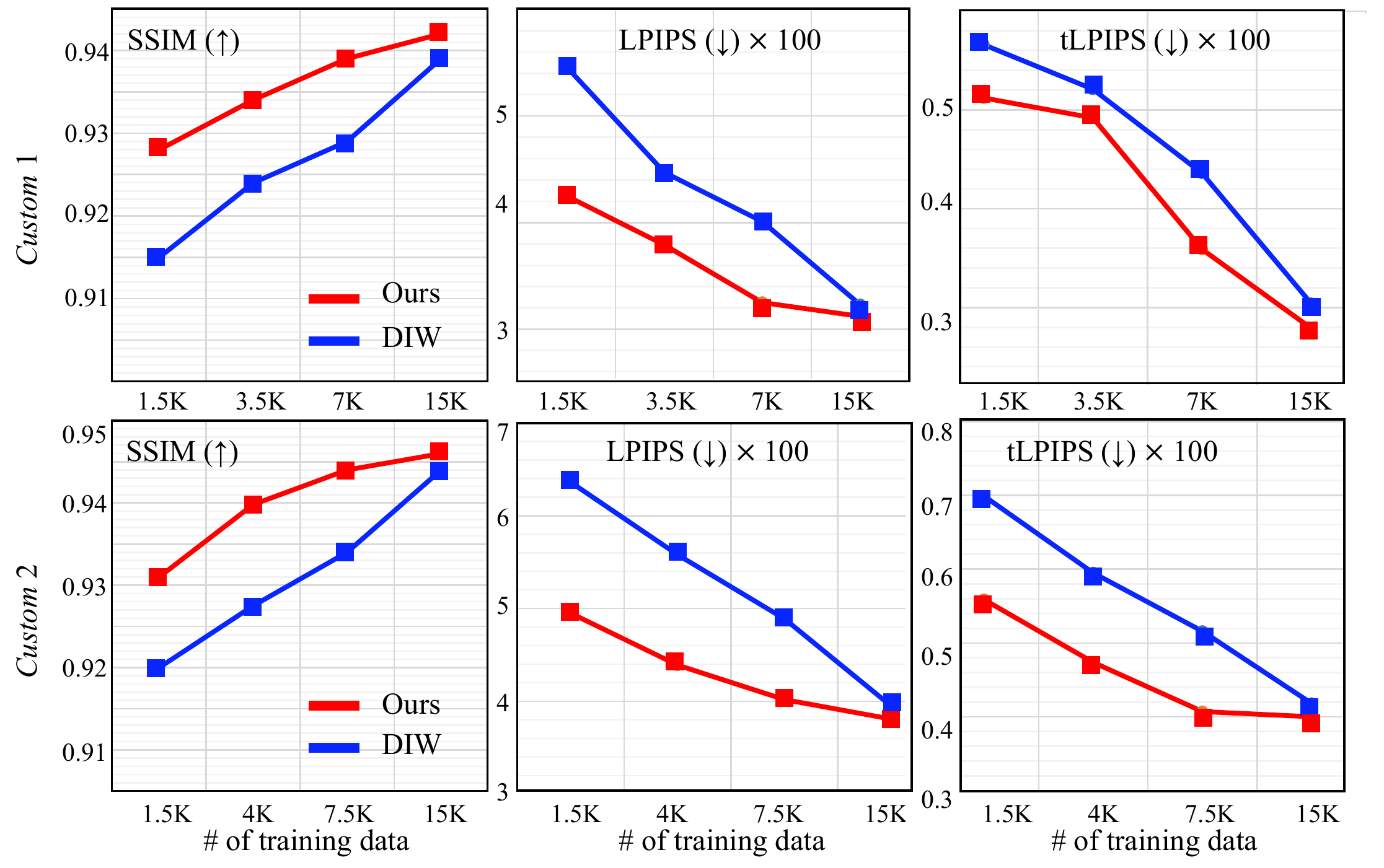}
	\end{center}	
	\vspace{-.4cm}
	\caption{\small Performance depending on the amount of training data. 
}
	\vspace{-.2cm}
	\label{fig:dat_to_quality}
\end{figure}

\begin{figure*}[t]
	\begin{center}
	\vspace{-.3cm}
    \includegraphics[width=1\linewidth]{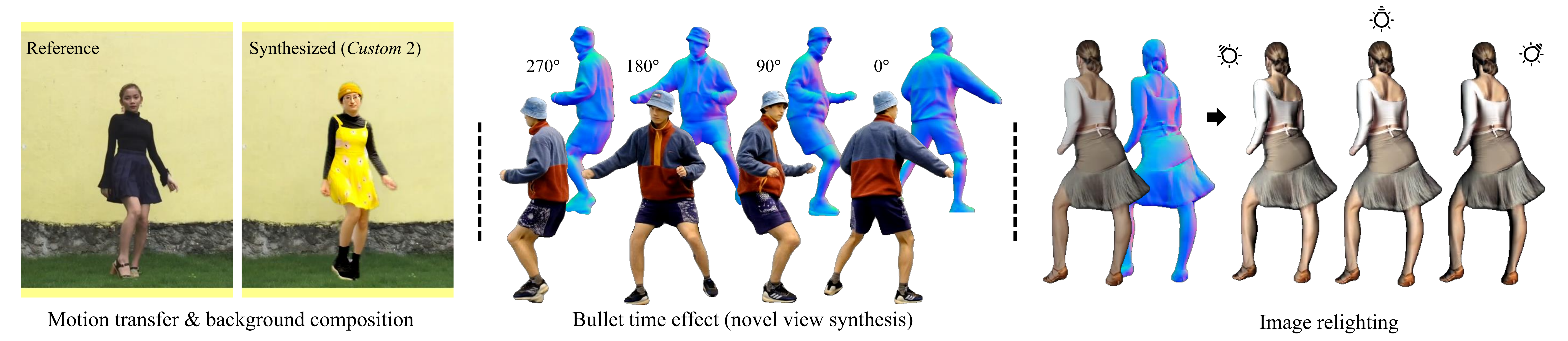}
	\end{center}	
	\vspace{-.6cm}
	\caption{\small Application. Our method enables several applications such as motion transfer with background composition, bullet time effects with novel view synthesis, and image-based relighting with the predicted surface normal.}
	\vspace{-.1cm}
	\label{fig:application}
\end{figure*}

{
\renewcommand{\tabcolsep}{5.5pt}
\begin{table*}[t]
\centering
\scriptsize
\begin{tabular}{l|cccccc|c}
\toprule
  Method   & \scriptsize{\textit{YouTube} 1 (0.6K)}  & \scriptsize{\textit{YouTube} 2 (1K)}  & \scriptsize{\textit{YouTube} 3 (0.4K)}  & \scriptsize{\textit{MPI} (1K)} & \scriptsize{\textit{Custom }1 (1.5K)}  & \scriptsize{\textit{Custom }2 (1.5K)} & \scriptsize{Avg.} \\
\hline
\scriptsize{DIW}~\cite{wang2021dance} &\scriptsize{0.939 / 4.12 / {\color{red}0.330}} &\scriptsize{0.940 / 5.00 / 0.463} &\scriptsize{0.869 / 6.75 / {\color{red}0.513}}   &\scriptsize{{\color{red}0.824} / 4.45 / 0.472} &\scriptsize{0.915 / 5.46 / 0.566}&  \scriptsize{0.920 / 6.36 / 0.698}&\scriptsize{0.901 / 5.36 / 0.507}      \\
\scriptsize{Ours} &\scriptsize{{\color{red}0.949} / {\color{red}3.36} / {0.457}} &\scriptsize{{\color{red}0.951} / {\color{red}4.16} / {\color{red}0.402}} &\scriptsize{{\color{red}0.883} / {\color{red}5.35} / 0.546}   &\scriptsize{0.824 / {\color{red}4.09} / {\color{red}0.327}} &\scriptsize{{\color{red}0.928} / {\color{red}4.25} / {\color{red}0.512}}&  \scriptsize{{\color{red}0.931} / {\color{red}4.95} / {\color{red}0.558}}&\scriptsize{{\color{red}0.911} / {\color{red}4.36} / {\color{red}0.467}}      \\
\bottomrule

\end{tabular}
\vspace{-1mm}
\caption{We train DIW~\cite{wang2021dance} and our method on a reduced training set (10$\%$ of the original training set) and test on the same testing set as in Table~\ref{table:full_res}. The thee numbers in each box represent the SSIM ($\uparrow$), LPIPS ($\downarrow$)$\times$100, and tLPIPS ($\downarrow$)$\times$100 metrics, respectively.
}
\label{table:body_experiment}
\end{table*}
}

\subsection{Evaluation}\label{sec:eval}

\noindent\textbf{Comparisons} 
We provide quantitative evaluation in Table~\ref{table:full_res} and show qualitative results in Figure~\ref{fig:render_res} (see Supplementary Video). Similar to our method, we train each baseline for roughly 72 hours until convergence. Both qualitative and quantitative results show that sparse 2D keypoint based pose representation used in EDN is not as effective as other baselines or our method. HFMT is successful in modeling dynamic appearance changes for mostly planar motions (i.e., MPI sequence), but shows inferior performance in remaining sequences that involve 3D rotations. This is due to the depth ambiguity inherent in sparse 2D keypoint based representation. While V2V performs well in terms of quantitative numbers, it suffers from significant texture drifting issues as shown in Figure~\ref{fig:render_res}, second row. We speculate that this is due to the errors in the optical flow estimation, especially in case of loose clothing, which is used as a supervisory signal. DIW uses dense UV coordinates to model the dynamic appearance changes of loose garments and is the strongest baseline. While it performs consistently well, we observe that the performance gap between DIW and our method increases for motion segments consisting of 3D rotations. This gap is magnified when the testing motion deviates from the training data. In Figure~\ref{fig:generalization_graph}, we plot how the perceptual error changes along the motion similarity between the training and testing data which is computed by NCC between two sets of the time-varying 3D meshes similar to Figure~\ref{fig:motivation}. We observe bigger increase in the error for DIW as testing frames deviate more from the training data. 


We next perform further comparisons with DIW evaluating the generalization ability of each approach.







%

\noindent\textbf{Generalization} 
An effective motion representation that encodes the dynamic appearance change of dressed humans should be discriminative to distinguish all possible deformations induced by a pose transformation given the current state of the body and the garments. In order to compare the discriminative power of our motion descriptor and the dense keypoint based representation proposed by DIW, we evaluate how well each representation generalizes to unseen poses. Specifically, we train each model using only $10\%$ of the original training sequences by subsampling the training frames while ensuring training and testing pose sequences are sufficiently distinct. Considering the reduced amount of data, we limit the training time to 24 hours for both approaches. As shown in Table~\ref{table:body_experiment}, the performance gap between the two methods increases. For the \textit{Custom} 1 and 2 sequences, we further repeat the same experiment using $10\%$, $25\%$, and $50\%$ of the original training data as shown in Figure~\ref{fig:dat_to_quality} where the performance of our method shows slower degradation than that of DIW. These quantitative results as well as visual results provided in the supplementary materials demonstrate the superiority of our 3D motion descriptor in terms of generalizing to novel poses. 

%

\noindent\textbf{Ablation Study} 
Using the \textit{Custom} 2 sequence, we train a variant \textbf{w/o 3D motion descriptors} by providing dense uv renderings as input directly to the decoder. We also disable the shape (\textbf{w/o shape}) and surface normal (\textbf{w/o surface normal}) prediction components. We repeat these trainings with subsampled data ($10\%$). As shown in Table~\ref{table:ablation_study}, the use of 3D motion descriptors and compositional rendering improves the perceptual quality of the synthesized images. The performance gap between our full model and w/o surface normal is larger with limited training data, implying that our multi-task framework helps with generalization. Qualitative results are given in the supplementary materials.

{
\renewcommand{\tabcolsep}{7pt}
\begin{table}[t]
\centering
\footnotesize
\begin{tabular}{l|cc}
\toprule
Method &  Full data (15K) & 10$\%$ data (1.5K) \\
\hline
w/o shape & \scriptsize{0.945 / 4.31 / {\color{red}0.401}} & \scriptsize{0.929 / 5.28/ 0.565}  \\
w/o surface normal & \scriptsize{ 0.945 / 3.89 / 0.418} & \scriptsize{0.929 / 5.17 / 0.602}  \\ 

w/o 3D motion & \scriptsize{ 0.942 / 4.17 / 0.584} & \scriptsize{0.928 / 5.43 / 0.760}  \\ 

\hline
Full & \scriptsize{{\color{red}0.946} / {\color{red}3.81} / 0.404} & \scriptsize{{\color{red}0.931} / {\color{red}4.95} / {\color{red}0.558}} \\ 
\bottomrule
\end{tabular}
\caption{Ablation study. The three metrics are SSIM ($\uparrow$), LPIPS ($\downarrow$)$\times$100, and tLPIPS ($\downarrow$)$\times$100 respectively. The number in the top row denotes the amount of training data. }
\label{table:ablation_study}
\end{table}
}



\subsection{Applications}\label{eval:app} 
Our method enables several additional applications as shown in Figure~\ref{fig:application}. Since our method works with 3D body based motion representation, it can be easily used to transfer motion from a source to a target character by simply transferring the joint rotations between the characters. We can also create bullet time effects by creating a target motion sequence by globally rotating the 3D body. Thanks to the surface normal prediction, we can also perform relighting which is otherwise not applicable. Please refer to the supplementary material for more details and results.

\section{Conclusion}\label{sec:conc}
We presented a method to render the dynamic appearance of a dressed human given a reference monocular video. Our method utilizes a novel 3D motion descriptor that encodes the time varying appearance of garments to model effects such as secondary motion. Our experiments show that our 3D motion descriptor is effective in modeling complex motion sequences involving 3D rotations. Our descriptor also demonstrates superior discriminator power compared to state-of-the-art alternatives enabling our method to better generalize to novel poses.

While showing impressive results, our method still has limitations. Highly articulated hand regions can appear blurry, hence refining the appearance of such regions with specialized modules is a promising direction. Our current model is subject specific, extending different parts of the model, e.g., 3D motion descriptor learning, to be universal is also an interesting future direction.


\noindent\textbf{Broader Impact} While our goal is to enable content creation for tasks such as video based motion retargeting or social presence, there can be some misuse of our technology to fabricate fake videos or news. We hope that parallel advances in deep face detection and image forensics can help to mitigate such concerns.

\section*{Acknowledgement}
We would like to thank Julien Philip for providing useful feedback on our paper draft. Jae Shin Yoon is supported by Doctoral Dissertation Fellowship from University of Minnesota. This work is partially supported by NSF CNS-1919965. 

{\small
\bibliographystyle{ieee_fullname}
\bibliography{main_arxiv.bbl}
}

\clearpage
\setcounter{section}{0}
\def\thesection{\Alph{section}}
\renewcommand{\thesubsection}{\thesection.\arabic{subsection}}

In the supplementary materials, we provide the details and evaluations of our 3D performance tracking method, network designs for our human rendering models, and more results and comparison. Please also refer to the supplementary video.

\section{Model-based Monocular 3D Pose Tracking}\label{method:perform}
\begin{figure}
	\begin{center}
     \includegraphics[width=1\linewidth]{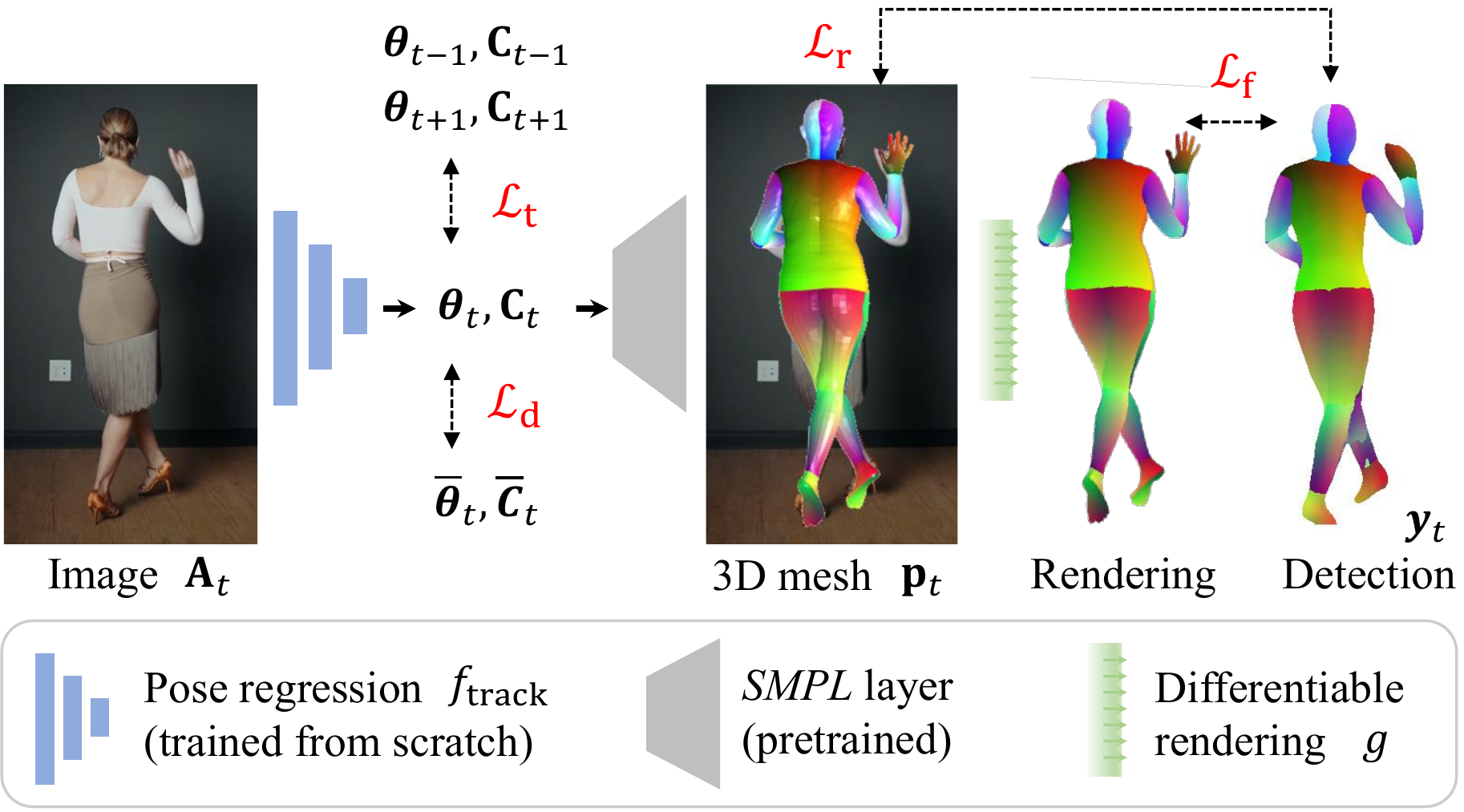}
	\end{center}	
	\vspace{-.4cm}
	\caption{\small 
The overview of our model-based monocular 3D performance tracking. A regression network predicts the body ($\boldsymbol{\theta}$) and camera ($\mathbf{C}$) pose parameters from a single image. The pretrained SMPL layer~\cite{SMPL:2015} decodes the predicted parameters to reconstruct the posed 3D body mesh. We render out the dense IUV coordinates of the mesh using a differentiable rendering layer and train the regression network by enforcing self-consistency between densepose detection and rendered IUV map~\cite{guler2018densepose} ($\mathcal{L}_{\rm r}$ and $\mathcal{L}_{\rm f}$); and enforcing temporal smoothness ($\mathcal{L}_{\rm t}$) and data-driven regularization ($\mathbf{L}_{\rm d}$).}
	\vspace{-.4cm}
	\label{fig:motion_caputre}
\end{figure}

Given a video of a moving person, we represent $\mathbf{p}$ as the posed 3D body at each frame. Specifically, we predict the parameters of the template SMPL model~\cite{SMPL:2015}, i.e., $\mathbf{p} = SMPL(\boldsymbol{\theta},\boldsymbol{\beta})$, where $SMPL$ is a function that takes the pose $\boldsymbol{\theta}\in\mathds{R}^{72}$ and shape $\boldsymbol{\beta}\in\mathds{R}^{10}$ parameters and provides the vertex locations of the 3D posed body. 
%
%
To this end, we learn a tracking function that regresses accurate and temporally coherent pose and camera parameters from an image sequence:
\begin{align}
\boldsymbol{\theta}_{t},\mathbf{C}_{t} =  f_{\rm track}(\mathbf{A}_{t}),\label{eq3_1}
\end{align}
where $f_{\rm track}$ is the tracking function, $\mathbf{A}_{t}$ is the image at time $t$, and $\mathbf{C}_{t}\in\mathds{R}^{3}$ is the camera translation relative to the body, camera rotation is encoded in $\boldsymbol{\theta}_t$. We assume the shape, $\boldsymbol{\beta}$, is constant. 
We use a weak-perspective camera projection model~\cite{kanazawa2018end} where we represent the camera translation in the z axis as the scale parameter. $f_{\rm track}$ is learned by minimizing the following loss for each input video:
\begin{align}
\mathcal{L}_{\rm track}= \mathcal{L}_{\rm f} + \lambda_{\rm r}\mathcal{L}_{\rm r} + \lambda_{\rm d}\mathcal{L}_{\rm d}+\lambda_{\rm t}\mathcal{L}_{\rm t},\label{eq3_1}
\end{align}
where $\mathcal{L}_{\rm f}$, $\mathcal{L}_{\rm r}$, $\mathcal{L}_{\rm d}$, and $\mathcal{L}_{\rm t}$ are the fitting, rendering, data prior, and temporal consistency losses, respectively, and $\lambda_{\rm r}$, $\lambda_{\rm d}$, and $\lambda_{\rm t}$ are their weights. We set $\lambda_{\rm r}=1, \lambda_{\rm d}=0.1$, and $\lambda_{\rm t}=0.01$ in our experiments. The overview of our optimization framework is described in Figure~\ref{fig:motion_caputre}.

$L_{\rm f}$ and $\mathcal{L}_{\rm r}$ utilize image-based dense UV map predictions~\cite{guler2018densepose} which enforce the 3D body fits to better align with the image space silhouettes of the body. Specifically, $L_{\rm f}$ measures the 2D distance between the projected 3D vertex locations and corresponding 2D points in the image:
\begin{align}
\mathcal{L}_{\rm f}=\sum_{\mathbf{X}\leftrightarrow \mathbf{x} \in \mathcal{U}}\|\Pi_p\mathbf{X} - \mathbf{x}\|.\label{eq5}
\end{align}
where $\mathcal{U}$ is the set of dense keypoints in the image, $\mathbf{x}\in\mathds{R}^2$ obtained from image-based dense UV map predictions~\cite{Yuanlu2019},  $\mathbf{X}$ are the corresponding 3D vertices, and $\Pi_p$ is the camera projection which is a function of $\mathbf{C}$.
%
$\mathcal{L}_{\rm r}$ measures the difference between the rendered and detected UV maps, $\mathbf{y}$: 
\begin{align}
\mathcal{L}_{\rm r}=\|g(\mathcal{W}^{-1}\mathbf{p}^{t},\mathbf{C}_{t})-\mathbf{y}\|,\label{eq7}
\end{align}
where $g(\cdot)$ is the differentiable rendering function that renders the UV coordinates from the 3D body model. 

$\mathcal{L}_{\rm d}$ provides the data driven prior on body and camera poses, i.e., $\mathcal{L}_{\rm d}=\|\boldsymbol{\theta}-\overline{\boldsymbol{\theta}}\|+\|\mathbf{C}-\overline{\mathbf{C}}\|$, where
$\overline{\boldsymbol{\theta}}$ and $\overline{\mathbf{C}}$ are the initial body and camera parameters predicted by a state-of-the-art method~\cite{kocabas2020vibe}. $\mathcal{L}_{\rm t}$ enforces the temporal smoothness over time: $\mathcal{L}_{\rm t}=\|\boldsymbol{\theta}_{t}-\boldsymbol{\theta}_{t-1}\|+ \|\boldsymbol{\theta}_{t}-\boldsymbol{\theta}_{t+1}\|+\|\mathbf{C}_{t}-\mathbf{C}_{t-1}\|+\|\mathbf{C}_{t}-\mathbf{C}_{t+1}\|$. 

\begin{figure}
	\begin{center}
    \includegraphics[width=0.9\linewidth]{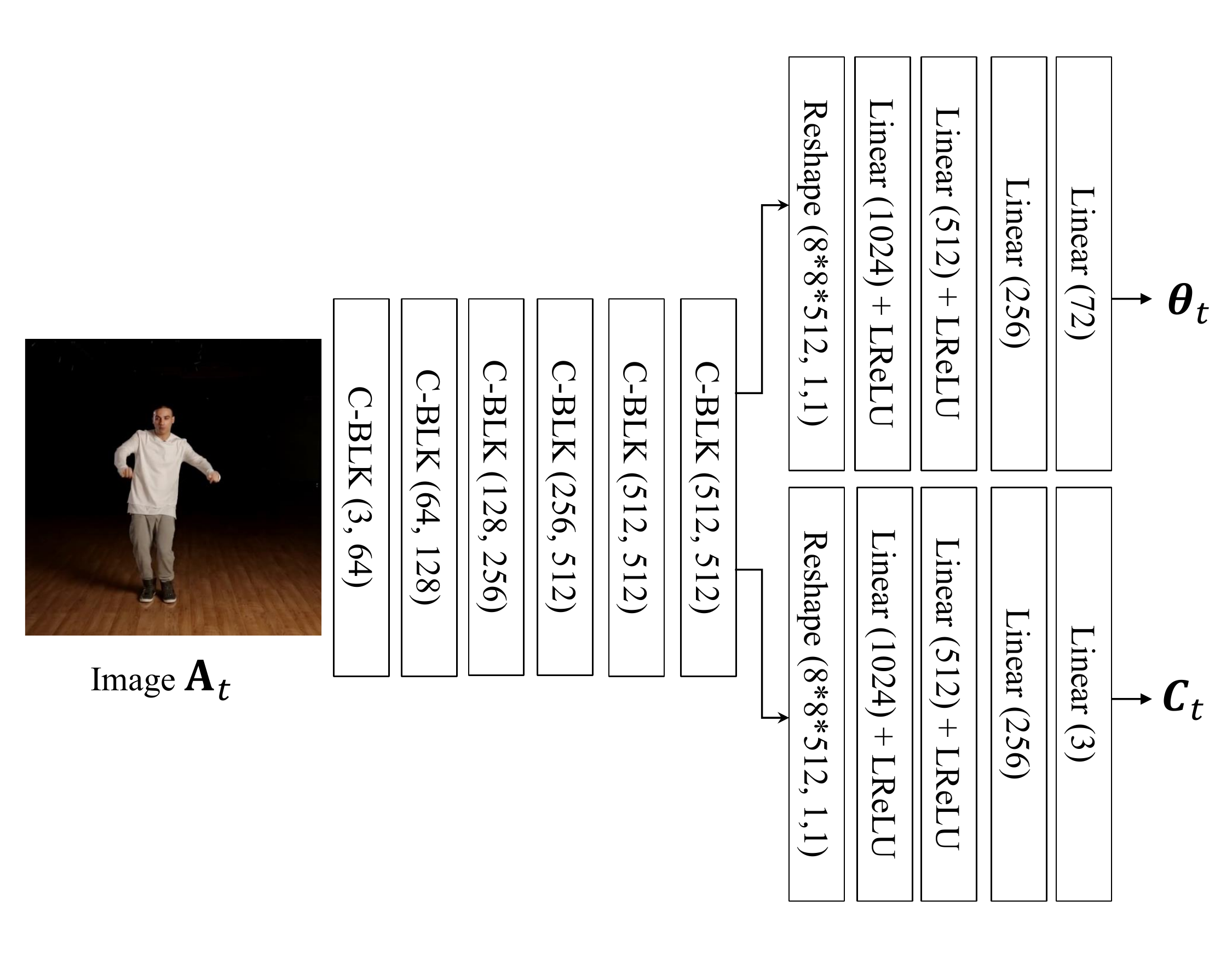}
	\end{center}	
	\vspace{-.8cm}
	\caption{\small Network design for our 3D body and camera pose regression network ($f_{\rm track}$). The details for C-BLK, D-BLK, Conv, and LReLU are described in Figure~\ref{fig:block}.}
	\vspace{-.4cm}
	\label{fig:regresser}
\end{figure}

We enable $f_{\rm track}$ using a convolutional neural network. The details of our network designs  are described in Figure~\ref{fig:regresser}. where it predicts the 3D body $\boldsymbol{\theta}$ and camera $\mathbf{C}$ pose from an image $\mathbf{A}$.



\begin{figure}
	\begin{center}
    \includegraphics[width=1\linewidth]{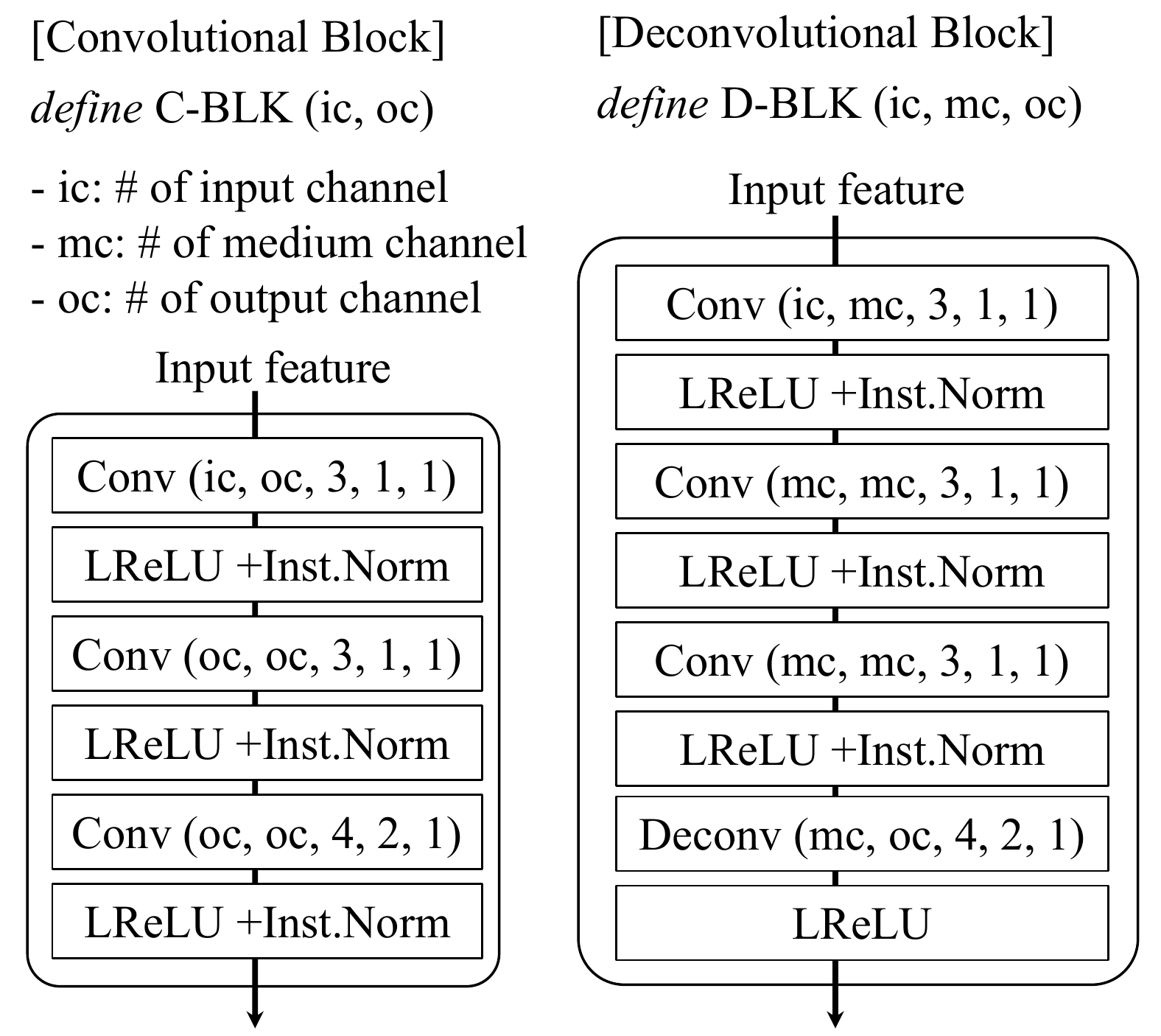}
	\end{center}	
	\vspace{-.4cm}
	\caption{\small Implementation details of our convolutional and deconvolutional blocks. Conv and Deconv denotes convolutional and deconvolutional layers are constructed based on the parameters: number of input channels (ic), number of output channels (oc), filter size, stride, and the size of the zero padding. We set the coefficient of the LeakyReLU (LReLU) to 0.2.}
	\label{fig:block}
\end{figure}

\noindent\textbf{Evaluation}\label{eval:pose}
We validate the performance of our 3D pose tracking method by comparing with previous monocular image based (SPIN~\cite{kolotouros2019learning}
and SMPLx~\cite{choutas2020monocular}) and video based (VIBE~\cite{kocabas2020vibe}) 3D body estimation methods.

\begin{figure}
	\begin{center}
    \includegraphics[width=1\linewidth]{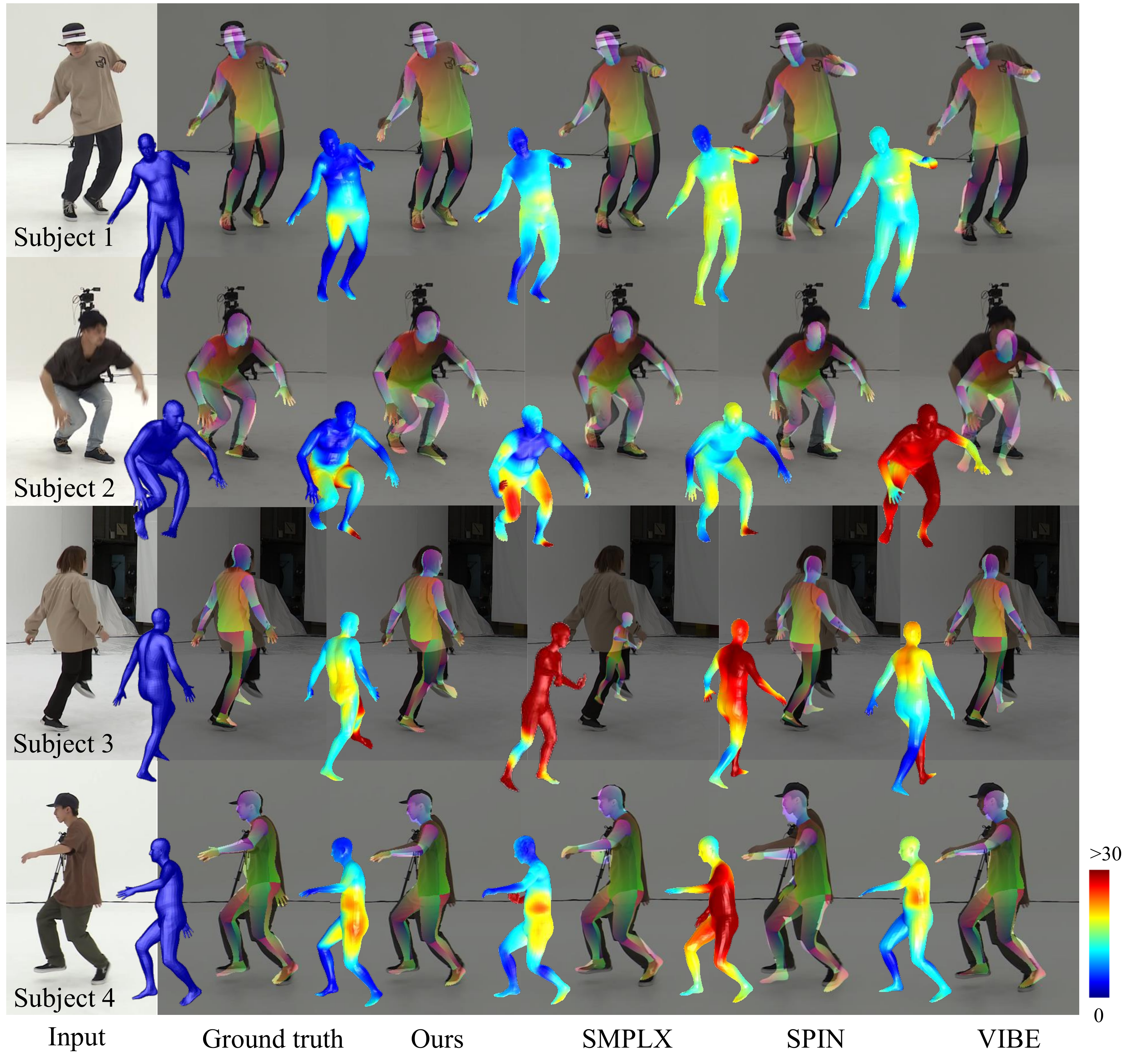}
	\end{center}	
	\vspace{-.4cm}
	\caption{\small We show the 3D body estimates and color coded 2D projection errors of our method and baselines for images of size $512 \times 512$.}
	\vspace{-.4cm}
	\label{fig:motion_result}
\end{figure}

We use the AIST++ dataset~\cite{li2021learn} which provides pseudo-ground truth SMPL fits obtained from multiview images. For randomly selected four subjects, we select four viewpoints and two motion styles (600 frames per motion) resulting in 4800 testing frames per subject.
Due to the differences in the camera models adopted by each method (\textit{i.e.,} perspective or orthographic cameras), there exist a scale ambiguity between the predictions and the ground truth. Hence, we measure the per-vertex 2D projection error between the ground truth and predicted 3D body model in the image space. We provide quantitative and qualitative results in Table~\ref{table:perfor_capture} and Figure~\ref{fig:motion_result}, respectively. By exploiting both temporal cues and dense keypoint estimates, our method outperforms the previous work.

{
\renewcommand{\tabcolsep}{3pt}
\begin{table}[t]
\centering
\footnotesize
\begin{tabular}{|l|c|c|c|c|c|c|}
\hline
&  Sub.1 & Sub.2 & Sub.3 & Sub.4 & Avg. \\
\hline
\scriptsize SPIN~\cite{kolotouros2019learning} & \scriptsize{16.5$\pm$3.7} & \scriptsize{22.6$\pm$6.6} & \scriptsize{23.4$\pm$6.2} &\scriptsize{ 21.5$\pm$4.4} & \scriptsize{21.0$\pm$5.2} \\
\hline
VIBE~\cite{kocabas2020vibe} & \scriptsize{13.9$\pm$2.9} & \scriptsize{12.2$\pm$2.8} & \scriptsize{17.7$\pm$5.1}  & \scriptsize{15.5$\pm$2.9}& \scriptsize{14.8$\pm$3.4} \\ 
\hline
SMPLx~\cite{choutas2020monocular} & \scriptsize{9.0$\pm$1.6} & \scriptsize{10.2$\pm$1.7} & \scriptsize{16.2$\pm$10.2} & \scriptsize{12.1$\pm$4.4} & \scriptsize{11.9$\pm$4.5}  \\ 
\hline
Ours & \scriptsize{{\color{red}8.3$\pm$1.1}} & \scriptsize{{\color{red}8.7$\pm$2.0}} & \scriptsize{{\color{red}13.7$\pm$3.5}}   & \scriptsize{{\color{red}11.3$\pm$1.9}}& \scriptsize{{\color{red}10.5$\pm$2.1}}\\ 
\hline

\end{tabular}
\caption{We show the mean and std of per-vertex projection error between the ground truth and estimated 3D bodies for images of size $512 \times 512$.}
\label{table:perfor_capture}
\end{table}
}

\section{Comparison to 3D based Approach}\label{more_comp}
\begin{figure}[h]
    \centering
    \includegraphics[width=1\linewidth]{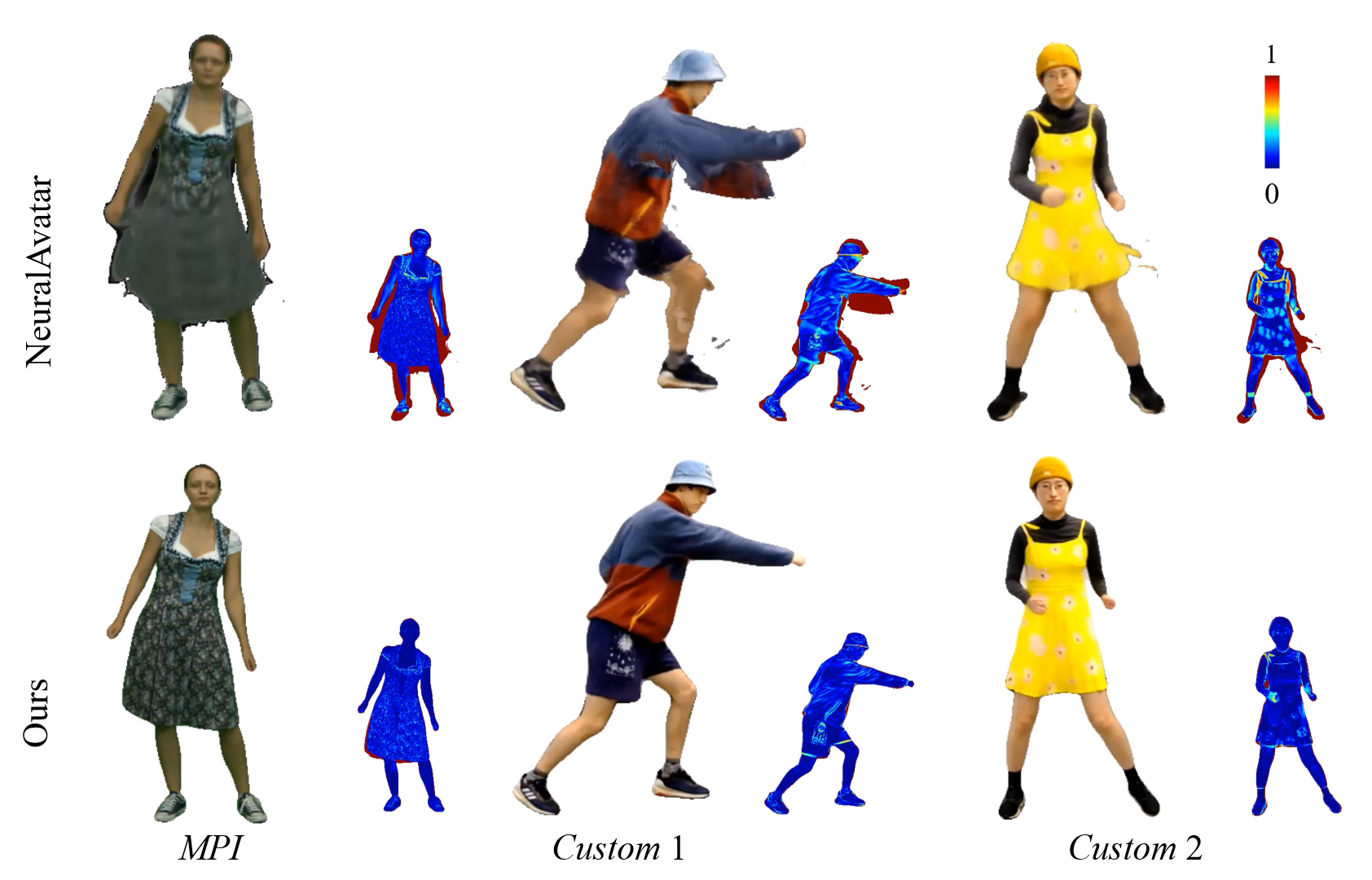}
    \caption{Qualitative comparison to NeuralAvatar~\cite{chen2021animatable}. The color map represents the per-pixel difference from real images.}
    \label{fig:comp}
\end{figure}

In Fig.~\ref{fig:comp} and Table~\ref{table:comp}, we show qualitative and quantitative comparisons to the recent 3D based method~\cite{chen2021animatable} for neural avatar modeling from a single camera, which explicitly reconstruct the geometry of animatable human. This method cannot effectively produce motion-dependent texture due to the failure in modeling of 3D deformation for clothing geometry, leading to rendering results with blurry, noisy, and static appearance as shown in Fig.~\ref{fig:comp}, upper row.

{
\renewcommand{\tabcolsep}{7pt}
\begin{table}[h]
\centering
\footnotesize
\begin{tabular}{l|ccc}
\toprule
 &  {\textit{MPI}} &{\textit{Custom} 1} &{\textit{Custom} 2}  \\
\hline
{NeuralAvatar~\cite{chen2021animatable}} & { 0.808 / 15.3} & {0.860 / 12.2} & {0.869 / 12.7}  \\ 

{Ours} & { \textbf{0.825} / \textbf{2.82}} & {\textbf{0.942} / \textbf{3.12}} &{\textbf{0.946} / \textbf{3.81}}    \\ 

\bottomrule
\end{tabular}
\vspace{-2mm}
\caption{\small Comparison to the 3D method. Two metrics represent SSIM ($\uparrow$), LPIPS ($\downarrow$)$\times$100, respectively. Three datasets are used. }
\vspace{-3mm}
\label{table:comp}
\end{table}
}


\section{More results}\label{more_res}
\begin{figure}
	\begin{center}
    \includegraphics[width=1.02\linewidth]{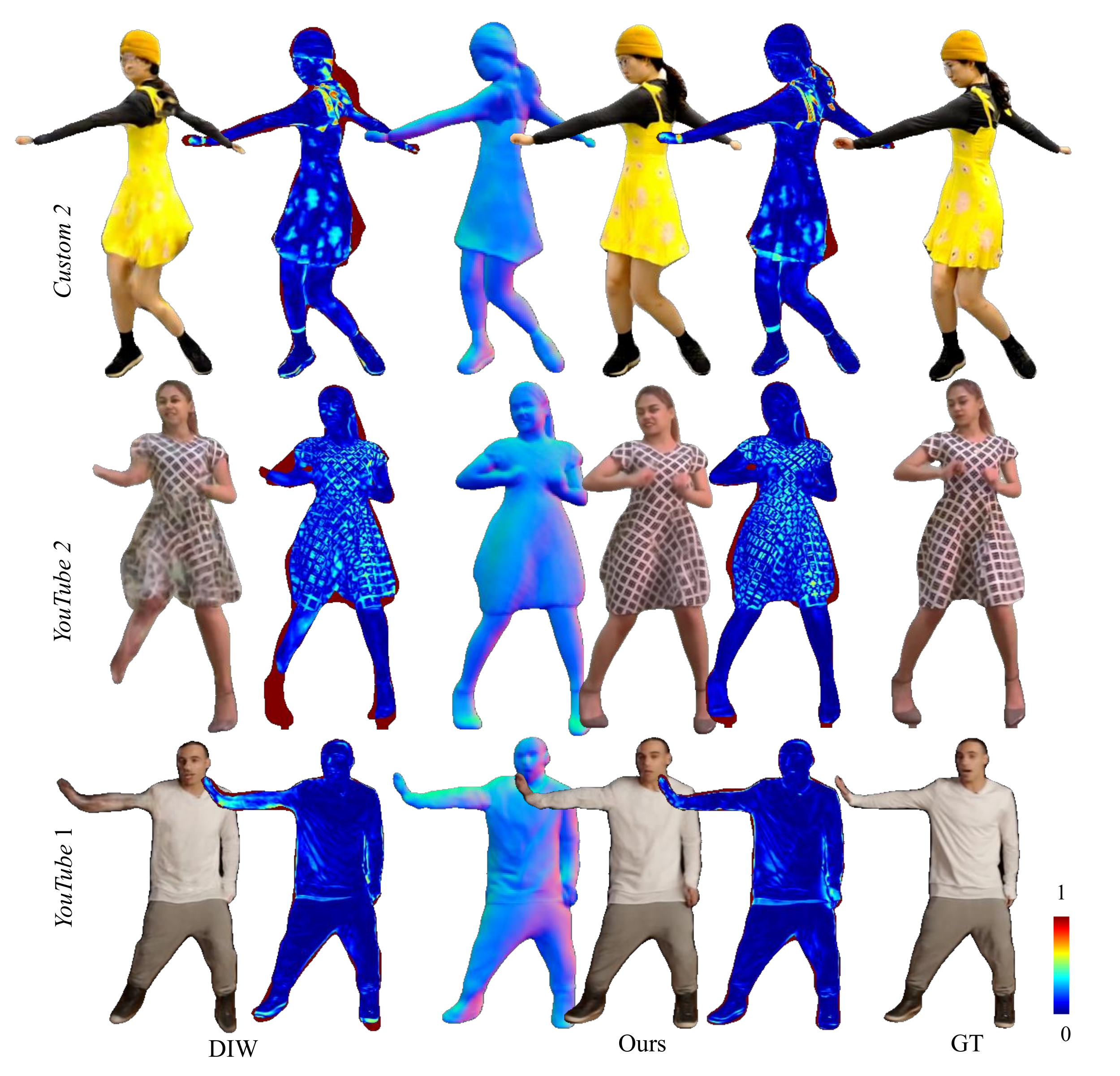}
	\end{center}	
	\vspace{-.4cm}
	\caption{\small Results from the model that learns from a small amount of data (10 $\%$ of full training data). The color map shows the pixel-wise difference of the synthesized image with the ground truth.}
	\vspace{-.4cm}
	\label{fig:render_res_small}
\end{figure}

In Figure~\ref{fig:render_res_small}, we show the qualitative results of our rendering model and the one from DIW~\cite{wang2021dance} which are trained with a small number of data (10$\%$ of the full training data). This result shows a similar trend as the Table~2 in the main manuscript.

\section{Network Designs for Human Rendering}\label{more_res}
We learn our motion encoder $E_{\Delta}$ and compositional rendering decoders, $E_{s}, E_{a}$ using convolutional neural networks. In this section, we provide the implementation details of our network designs.

\begin{figure*}
	\begin{center}
    \includegraphics[width=0.8\linewidth]{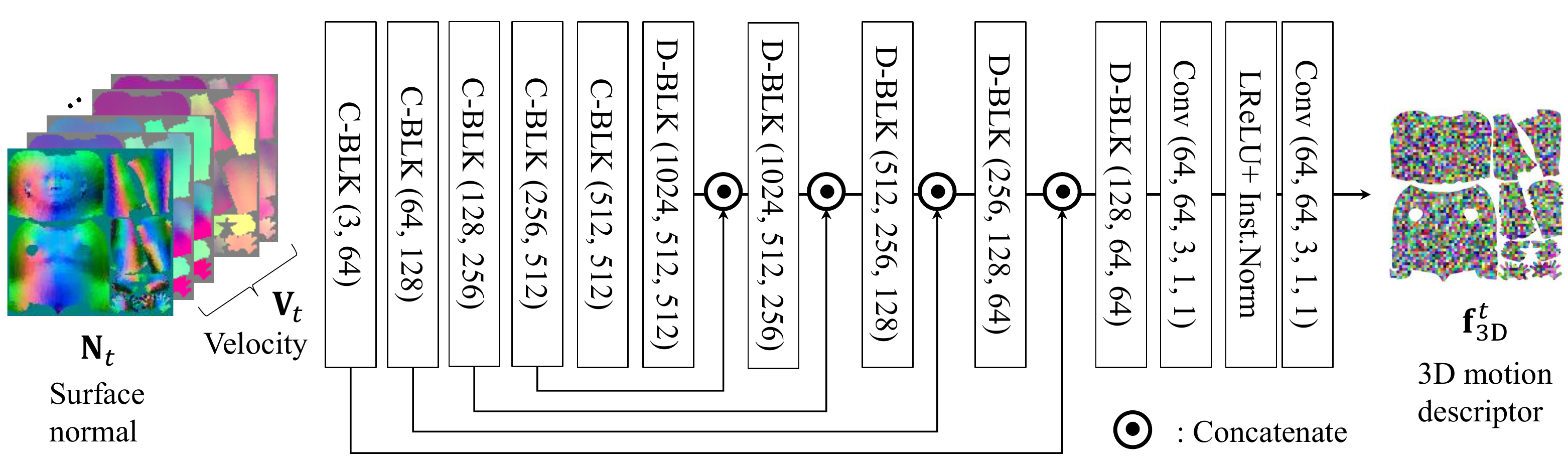}
	\end{center}	
	\vspace{-3mm}
	\caption{\small Network design for our 3D motion encoder ($E_{\Delta}$). The details of C-BLK, D-BLK, Conv, and LReLU are described in Figure~\ref{fig:block}.}
	\label{fig:motionnet}
\end{figure*}

\noindent\textbf{3D Motion Encoder Network, $E_{\Delta}$.} 
Figure~\ref{fig:motionnet} describes the network details for our 3D motion encoder $E_{\Delta}$. It takes as input 3D surface normal $\mathbf{N}_t$ of the current frame and velocity $\mathbf{V}_t$ for past $10$ frames recorded in the UV space of the body and outputs 3D motion descriptors $\mathbf{f}_{\rm 3D}^{t}$. 

\begin{figure*}
	\begin{center}
    \includegraphics[width=0.9\linewidth]{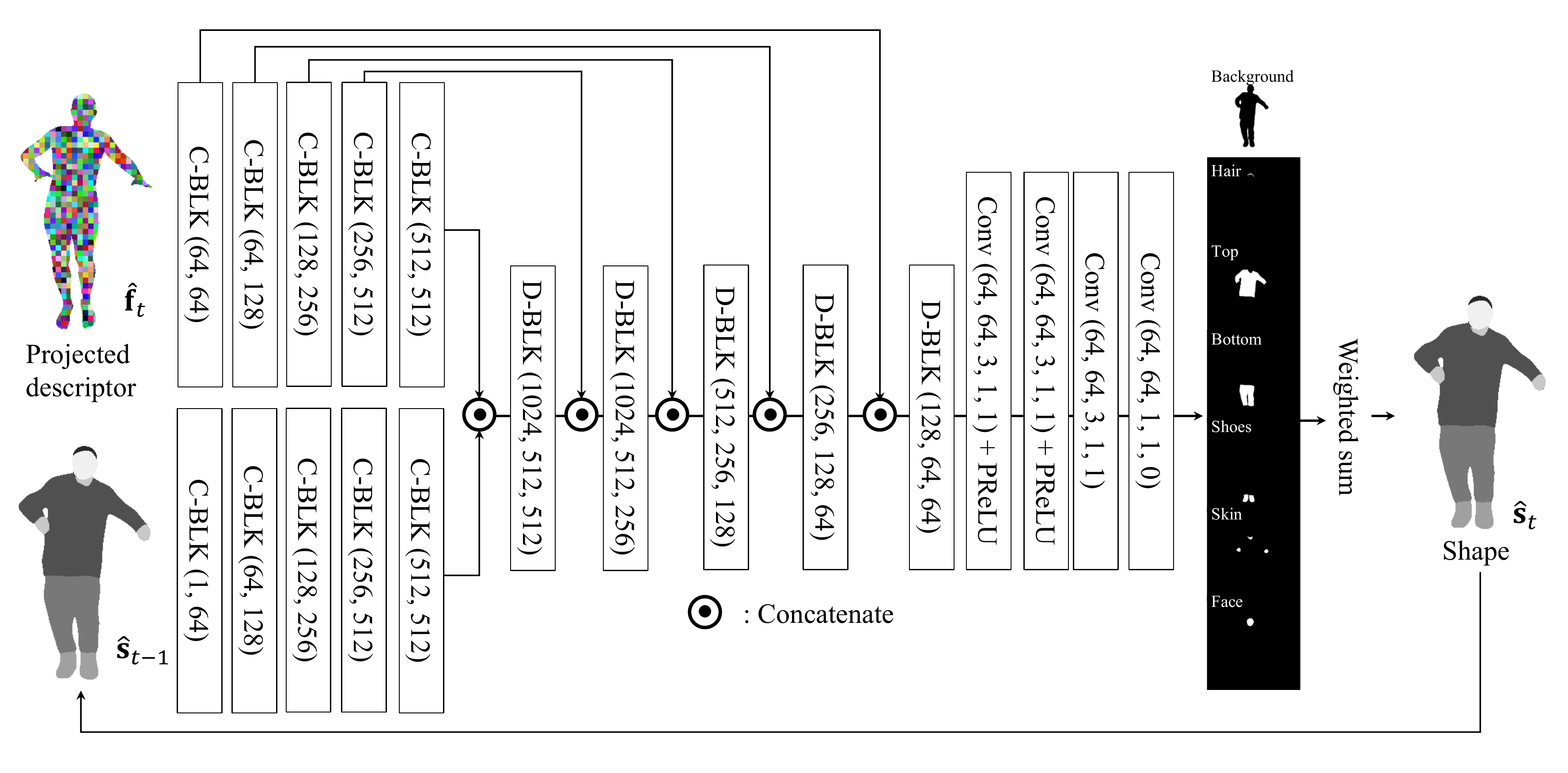}
	\end{center}	
	\vspace{-.8cm}
	\caption{\small Network design for our shape decoder ($D_{s}$). The details of C-BLK, D-BLK, Conv, and LReLU are described in Figure~\ref{fig:block}.}
	\vspace{-.4cm}
	\label{fig:shapenet}
\end{figure*}

\noindent\textbf{Shape Decoder Network, $E_{s}$.} 
Figure~\ref{fig:shapenet} describes the network details for our shape decoder network $E_{s}$ which takes as input the 3D motion descriptor $\widehat{f}_{t}$ rendered in the image space and the predicted shape in the previous time instance $\widehat{\mathbf{s}}_{t-1}$, and outputs the person-specific 2D shape $\widehat{\mathbf{s}}_{t}$ which is composed seven category label maps.

\begin{figure*}
	\begin{center}
    \includegraphics[width=0.9\linewidth]{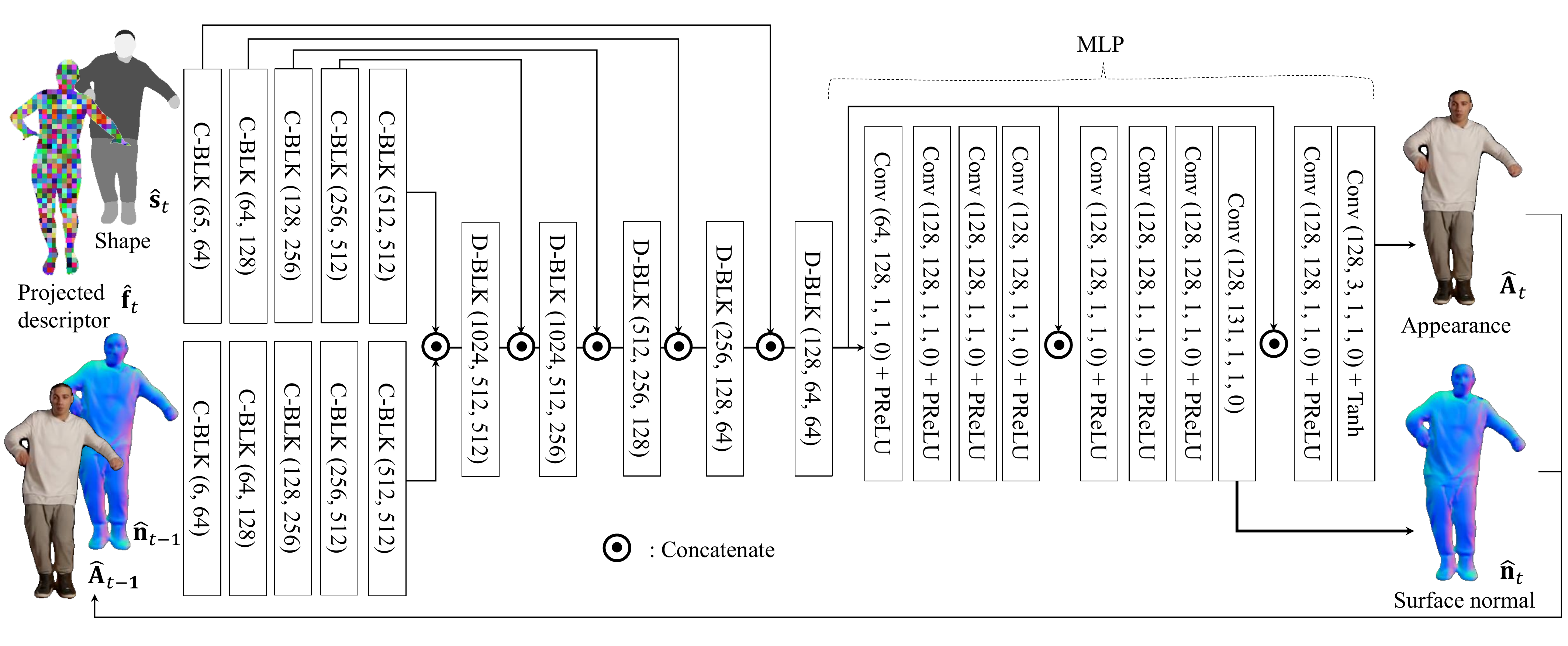}
	\end{center}	
	\vspace{-.8cm}
	\caption{\small Network design for our appearance decoder ($D_{a}$). The details of C-BLK, D-BLK, Conv, and LReLU are described in Figure~\ref{fig:block}.}
	\vspace{-.4cm}
	\label{fig:imagenet}
\end{figure*}

\noindent\textbf{Appearance Decoder Network, $E_{a}$.} 
Figure~\ref{fig:imagenet} describes the network details for our appearance decoder network $E_{s}$ which takes as input the projected 3D motion descriptor $\widehat{f}_{t}$ rendered in the image space, predicted shape $\widehat{\mathbf{s}}_{t}$, and the predicted appearance and surface normal in the previous time instance $\{\widehat{\mathbf{A}}_{t-1}, \widehat{\mathbf{n}}_{t-1}\}$, and outputs the 3D surface normal $\widehat{\mathbf{n}}_{t}$ and appearance $\widehat{\mathbf{A}}_{t}$.

\end{document}